\documentclass[conference]{IEEEtran}
\usepackage{times}

\usepackage[numbers]{natbib}
\usepackage{multicol}
\usepackage[bookmarks=true]{hyperref}

\usepackage[ruled,vlined]{algorithm2e}
\usepackage{amsmath}
\usepackage{amssymb}
\usepackage{graphicx}

\begin{document}

\title{Gaussian Process-Based Decentralized Data Fusion and Active Sensing for Mobility-on-Demand System}

\author{
\authorblockN{Jie Chen, Kian Hsiang Low, and Colin Keng-Yan Tan}
\authorblockA{Department of Computer Science, 
National University of Singapore, Republic of Singapore\\ 
\{chenjie, lowkh, ctank\}@comp.nus.edu.sg}
}



%

\newcommand{\set}[1]{{\mathcal #1}}
\newcommand{\squishlisttwo}{
 \begin{list}{$\bullet$}
  { \setlength{\itemsep}{0pt}
     \setlength{\parsep}{0pt}
    \setlength{\topsep}{0pt}
    \setlength{\partopsep}{0pt}
    \setlength{\leftmargin}{1em}
    \setlength{\labelwidth}{1.5em}
    \setlength{\labelsep}{0.5em} } }

\newcommand{\squishend}{
  \end{list}  }

\newtheorem{defi}{Definition}
\newtheorem{thm}{Theorem}
\def\myproof{1} 

\maketitle

\begin{abstract}
Mobility-on-demand (MoD) systems have recently emerged as a promising paradigm of one-way vehicle sharing for sustainable personal urban mobility in densely populated cities.
In this paper, we enhance the capability of a MoD system by deploying robotic shared vehicles that can autonomously cruise the streets to be hailed by users. A key challenge to managing the MoD system effectively is that of real-time, fine-grained mobility demand sensing and prediction.
This paper presents a novel decentralized data fusion and active sensing algorithm for real-time, fine-grained mobility demand sensing and prediction with a fleet of autonomous robotic vehicles in a MoD system. Our Gaussian process (GP)-based decentralized data fusion algorithm can achieve a fine balance between predictive power and time efficiency. We theoretically guarantee its predictive performance to be equivalent to that of a sophisticated centralized sparse approximation for the GP model: The computation of such a sparse approximate GP model can thus be distributed among the MoD vehicles, hence achieving efficient and scalable demand prediction. Though our decentralized active sensing strategy is devised to gather the most informative demand data for demand prediction, it can achieve a dual effect of fleet rebalancing to service the mobility demands. Empirical evaluation on real-world mobility demand data shows that our proposed algorithm can achieve a better balance between predictive accuracy and time efficiency than state-of-the-art algorithms. 
\end{abstract}

\IEEEpeerreviewmaketitle

\section{Introduction} \label{sec:intro} 
%
%
%
Private automobiles are becoming unsustainable personal mobility solutions in densely populated urban cities because the addition of parking and road spaces cannot keep pace with their escalating numbers due to limited urban land.
For example, Hong Kong and Singapore have, respectively, experienced $27.6\%$ and $37\%$ increase in private vehicles from $2003$ to $2011$
\cite{RPT}.
However, their road networks have only expanded less than $10\%$ in size.
Without implementing sustainable measures, traffic congestions and delays will grow more severe and frequent, especially during peak hours.

\emph{Mobility-on-demand} (MoD) systems \cite{Mitchell10} (e.g., V{\'{e}}lib system of over $20000$ shared bicycles in Paris, experimental car-sharing systems described in \cite{Frazzoli12}) have recently emerged as a promising paradigm of one-way vehicle sharing for sustainable personal urban mobility, specifically, to tackle the problems of low vehicle utilization rate and parking space caused by private automobiles. 
Conventionally, a MoD system provides stacks and racks of light electric vehicles distributed throughout a city: When a user wants to go somewhere, he simply walks to the nearest rack, swipes a card to pick up a vehicle, drives it to the rack nearest to his destination, and drops it off.
In this paper, we enhance the capability of a MoD system by deploying robotic shared vehicles (e.g., General Motors Chevrolet EN-V 2.0 prototype \cite{GM}) 
that can autonomously drive and cruise the streets of a densely populated urban city to be hailed by users (like taxis) instead of just waiting at the racks to be picked up. 
Compared to the conventional MoD system, the fleet of autonomous robotic vehicles 
provides greater accessibility to users who can be picked up and dropped off at any location in the road network. As a result, it can service regions of high mobility demand but with poor coverage of stacks and racks due to limited space for their installation.

The key factors in the success of a MoD system are the costs to the users and system latencies, which can be minimized by managing the MoD system effectively.
To achieve this, two main technical challenges need to be addressed \cite{Mitchell08}: (a) Real-time, fine-grained mobility demand sensing and prediction, and (b) real-time active fleet management to balance vehicle supply and demand and satisfy latency requirements at sustainable operating costs.
Existing works on load balancing for MoD systems \cite{Frazzoli12}, dynamic traffic assignment problems \cite{Peeta01}, dynamic one-to-one pickup and delivery problems \cite{Laporte10}, and location recommendation and dispatch for cruising taxis \cite{chang10,Ge10,Li12,Zheng12} have tackled variants of the second challenge by assuming the necessary input of mobility demand information to be perfectly or accurately known using prior knowledge or offline processing of historic data.
Such an assumption does not hold for densely populated urban cities because their mobility demand patterns are often subject to short-term random fluctuations and perturbations, in particular, due to frequent special events (e.g., storewide sales, exhibitions), unpredictable weather conditions, or emergencies (e.g., breakdowns in public transport services).
So, in order for the active fleet management strategies to perform well, they require accurate, fine-grained information of the spatiotemporally varying mobility demand patterns in real time, which is the desired outcome of addressing the first challenge.
To the best of our knowledge, there is little progress in the algorithmic development of the first challenge, which will be the focus of our work in this paper.

Given that the autonomous robotic vehicles are used directly as mobile probes to sample real-time demand data (e.g., pickup counts of different regions), 
how can the vacant ones actively cruise a road network to gather and assimilate the \emph{most informative} demand data for predicting the mobility demand pattern?
To solve this problem, a centralized approach to data fusion and active sensing \cite{LowAAMAS13,Guestrin08,LowAAMAS08,low09,LowAAMAS11,LowAAMAS12} is poorly suited because it suffers from a single point of failure and incurs huge communication, space, and time overheads with large data and fleet. 
%
Hence, we propose a novel decentralized data fusion and active sensing 
algorithm for real-time, fine-grained mobility demand sensing and prediction with a fleet of autonomous robotic vehicles in a MoD system.
The specific contributions of our work include: 
%
%
\squishlisttwo
\item Modeling and predicting the spatiotemporally varying mobility demand pattern using a rich class of Bayesian nonparametric models called the \emph{log-Gaussian process} ($\ell$GP)
(Section~\ref{sec:lgp}); 
%
\item Developing a novel \emph{Gaussian process-based decentralized data fusion} (GP-DDF$^+$) algorithm (Section~\ref{sec:ddf}) that achieves a fine balance between predictive power and time efficiency, and theoretically guaranteeing its predictive performance to be  equivalent to that of a sophisticated centralized sparse approximation for the Gaussian process (GP) model: The computation of such a sparse approximate GP model can thus be distributed among the MoD vehicles, hence achieving efficient and scalable demand prediction;
%
\item Exploiting a \emph{decentralized active sensing} (DAS) strategy (Section~\ref{sec:das}) to gather the most informative demand data for predicting the mobility demand pattern: Interestingly, we can analytically show that DAS exhibits a cruising behavior of simultaneously exploring demand hotspots and sparsely sampled regions that have higher likelihood of picking up users, hence achieving a dual effect of fleet rebalancing to service the mobility demands;
\item Empirically evaluating the predictive accuracy, time efficiency, scalability, and performance of servicing mobility demands (i.e., average cruising length of vehicles, average waiting time of users, total number of pickups) of our proposed algorithm on a real-world mobility demand pattern over the central business district of Singapore during evening hours (Section~\ref{sec:exp}).
%
\squishend
%
\section{Modeling a Mobility Demand Pattern} \label{sec:lgp} 
The service area in an urban city can be represented as a directed
graph $G\triangleq(V, E)$ where $V$ denotes a set of all regions generated by
gridding the service area, and $E\subseteq V\times V$ denotes a set of
edges such that there is an edge $(s,s')$ from $s\in V$ to $s'\in V$
iff at least one road segment in the road network starts in $s$ and ends in
$s'$.
Each region $s\in V$ is associated with a $p$-dimensional feature vector $x_s$ representing its context information (e.g., location, time, precipitation), and a measurement $y_s$ quantifying its mobility demand\footnote{The service area is represented as a grid of regions instead of a network of road segments like in \cite{chen12} because we observe less smoothly-varying, noisier demand measurements (hence, lower spatial correlation) for the latter in our real-world data (Section~\ref{sec:exp}) since many road segments do not permit stopping of vehicles.}.
Since it is often impractical in terms of sensing resource cost to determine the actual mobility demand of a region, a common practice is to use the pickup count of the region as a surrogate measure. 
To elaborate, the user pickups made by vacant MoD vehicles cruising in a region contribute to its pickup count. Since we do not assume a data center to be available to keep track of the pickup count, a fully distributed gossip-based protocol \cite{jelasity05} is utilized to aggregate these pickup information from the vehicles in the region that are connected via an ad hoc wireless communication network.
Consequently, any vehicle entering the region can access its pickup count simply by joining its ad hoc network.

As observed in \cite{chang10,Li12} and our real-world data (Fig.~\ref{fig:dmd}a), a mobility demand pattern over a large service area in an urban city is typically characterized by spatiotemporally correlated demand measurements and contains a few small-scale
hotspots exhibiting extreme measurements and much higher spatiotemporal variability than the rest of the demand pattern. That is,
if the measurements are put together into a $1$D sample frequency distribution, a positive skew results.
%
We like to consider using a rich class of Bayesian nonparametric models called \emph{Gaussian process} (GP) \cite{rasmussen06} to model the demand pattern.
But, the GP covariance structure is sensitive to strong positive skewness and easily destabilized by a few extreme measurements \cite{Webster01}.
In practice, this can cause reconstructed patterns to display large hotspots centered about a few extreme measurements and predictive variances to be unrealistically small in hotspots \cite{Hohn98}, which are undesirable.
So, if the GP is used to model a demand pattern directly, it may not predict well.
To resolve this, a standard statistical practice is to take the log of the measurements (i.e., $z_s=\log y_s$)
to remove skewness and extremity, and use the GP to model the demand pattern in the \emph{log-scale} instead.
%
\subsection{Gaussian Process (GP)} 
Each region $s\in{V}$ is associated with a realized (random) log-measurement $z_s$ ($Z_s$) if $s$ is sampled/observed (unobserved).
Let $\{Z_s\}_{s \in {V}}$ denote a GP, that is, every finite subset of $\{Z_s\}_{s \in {V}}$ has a multivariate Gaussian distribution.
The GP is fully specified by its \emph{prior} mean $\mu_s\triangleq \mathbb{E}[Z_s]$ and covariance $\sigma_{ss'}\triangleq\mbox{cov}(Z_s,Z_{s'})$ for all $s, s' \in V$,
the latter of which is defined by the widely-used squared exponential covariance function:
%
\begin{equation*}
\sigma_{ss'}\triangleq
\sigma_s^2\exp\left({-\frac{1}{2}\sum_{i=1}^{p}
\left(\frac{[x_s]_i-[x_{s'}]_i}{\ell_i}\right)^2}\right)+\sigma_n^2\delta_{ss'}
\label{eq:seard} 
\end{equation*}
where $[x_s]_i \left([x_{s'}]_i \right)$ is the $i$-th component of the feature vector $x_s \left(x_{s'}\right)$, the hyperparameters $\sigma^2_n,\sigma^2_s,\ell_1,\dots,\ell_p$ are, respectively, noise and signal variances and length-scales that can be learned using maximum likelihood estimation, and $\delta_{s s'}$ is a Kronecker delta that is
$1$ if $s = s'$ and $0$ otherwise. 
Given a set $D\subset V$ of observed regions and a column vector $z_D$ of corresponding log-measurements, the GP can be used to predict the log-measurements of any set $S\subset V$ of unobserved regions with the following Gaussian \emph{posterior} mean vector and covariance matrix:
\begin{equation}\vspace{-0mm}
\displaystyle\mu_{S|D}\triangleq \mu_{S} + \Sigma_{SD}\Sigma_{DD}^{-1}(
    z_D - \mu_D )
\label{eq:fgpmu} 
\end{equation}
\begin{equation}\vspace{-0mm}
\displaystyle\Sigma_{SS|D}\triangleq
\Sigma_{SS}-\Sigma_{SD}\Sigma_{DD}^{-1}\Sigma_{DS}
\label{eq:fgpcov} 
\end{equation}
where $\mu_S (\mu_D)$ is a column vector with mean components $\mu_s$
for all $s\in S (s\in D)$, $\Sigma_{SD}(\Sigma_{DD})$ is a covariance
matrix with covariance components $\sigma_{ss'}$ for all $s\in S, s'\in
D (s,s'\in D)$, and $\Sigma_{SD}$ is the transpose of $\Sigma_{DS}$.

\subsection{Log-Gaussian Process ($\ell$GP)} 
Demand measurements may not be observed in some regions because vacant MoD vehicles did not cruise into them.
Since our ultimate interest is to predict them in the \emph{original scale}, GP's predicted log-measurements of these unobserved regions must be transformed back \emph{unbiasedly}.
To achieve this, we utilize a widely-used variant of GP in geostatistics called the $\ell$GP that can model the demand pattern in the original scale.
Let $\{Y_s\}_{s\in V}$ denote a $\ell$GP: If $Z_s\triangleq \log Y_s$, then $\{Z_s\}_{s\in V}$ is a GP. 
So, $Y_s = \exp\{Z_s\}$ denotes the original random demand measurement of unobserved region $s$ and is predicted using the log-Gaussian posterior mean (i.e., best unbiased predictor)
\begin{equation}
\widehat{\mu}_{s|D}\triangleq\exp(\mu_{s|D}+{\Sigma_{ss|D}}/2)  
\label{eq:lgpmu}
\end{equation}
where $\mu_{s|D}$ and $\Sigma_{ss|D}$ are simply the Gaussian posterior mean (\ref{eq:fgpmu}) and variance (\ref{eq:fgpcov}) of GP, respectively.
 %
%
 %
%
%
%
The uncertainty of predicting the measurements of any set $S\subset V$ of unobserved regions can be quantified by the following log-Gaussian posterior joint entropy, which will be exploited by our DAS strategy (Section~\ref{sec:das}): 
\begin{equation}\vspace{-0mm}
\mathbb{H}[Y_S|Y_D]\triangleq
\frac{1}{2}\log\hspace{-0.5mm}\left(2\pi e\right)^{\left|S\right|}
\left|\Sigma_{SS|D}\right| +  \mu_{S|D}\cdot\mathbf{1} 
\label{eq:lgpent} \end{equation}
where $\mu_{S|D}$ and $\Sigma_{SS|D}$ are the Gaussian posterior mean vector (\ref{eq:fgpmu}) and covariance matrix (\ref{eq:fgpcov}) of GP, respectively.
%
%

%
\section{Decentralized Demand Data Fusion} \label{sec:ddf}
%
%
The demand data are gathered in a distributed manner by the vacant MoD vehicles cruising the service area and have to be assimilated in order to predict the mobility demand pattern.
A straightforward approach to data fusion is to fully communicate all the data to every vehicle, each of which then performs the same exact $\ell$GP prediction (\ref{eq:lgpmu}) separately.
This approach, which we call full Gaussian process (FGP) \cite{LowAAMAS08,low09},
unfortunately cannot scale well and be performed in real time due to its cubic time complexity in the size of the data.
%

Alternatively, the work of
\cite{chen12} has recently proposed a Gaussian process-based decentralized data fusion (GP-DDF)
  approach to efficient and scalable approximate GP and $\ell$GP
  prediction. Their key idea is as follows: Each vehicle summarizes all its local data, based on a common prior support set $U\subset V$, into a local summary (Definition~\ref{def:ls}) to be exchanged with every other vehicle. Then, it assimilates the local summaries received from the other vehicles into a global summary (Definition~\ref{def:gs}), which is then exploited for predicting the demands of unobserved regions. 
Though GP-DDF scales very well with large data, it can predict poorly due to (a) 
loss of information caused by summarizing the measurements and correlation
structure of the original data; and (b) sparse coverage of the hotspots (i.e., with higher spatiotemporal variability) by the support set.

We propose a novel decentralized data fusion algorithm called GP-DDF$^+$ that combines the best of both worlds, that is, the predictive power of FGP and efficiency of GP-DDF.
GP-DDF$^+$ is based on the intuition that a vehicle can exploit its local data to improve the demand predictions for unobserved regions ``close'' to its data (in the correlation sense).
At the same time, GP-DDF$^+$ can preserve the efficiency of GP-DDF by exploiting its idea of summarizing information, specifically, into the local and global summaries, as reproduced below:\vspace{2mm}

\begin{defi}[Local Summary] Given a common support set $U\subset V$
known to all $K$ vehicles, a set $D_k \subset V$ of observed regions 
 and a column vector $z_{D_k}$  of corresponding demand measurements
local to vehicle $k$, its local summary is defined as a tuple
$(\dot{z}_U^k, \dot{\Sigma}_{UU}^k)$ where
\begin{equation}
\dot{z}_{B}^k \triangleq \Sigma_{BD_k}\Sigma^{-1}_{D_k
D_k|U}(z_{D_k}-\mu_{D_k})
\label{eq:lsf} \end{equation}
\begin{equation}
\dot{\Sigma}_{BB'}^k\triangleq
\Sigma_{BD_k}\Sigma_{D_k D_k|U}^{-1}\Sigma_{D_k B'}
\label{eq:lsc}
\end{equation}
such that $\Sigma_{D_k D_k|U}$ is defined in a similar manner as
(\ref{eq:fgpcov}). 
\label{def:ls}\vspace{-0mm} 
\end{defi}
\vspace{2mm}

\begin{defi}[Global Summary] Given a common support set $U\subset V$
known to all $K$ vehicles and the local summary $(\dot{z}_U^k,
    \dot{\Sigma}_{UU}^k)$ of every vehicle $k = 1,\ldots,K$, the global
summary is defined as a tuple $(\ddot{z}_U, \ddot{\Sigma}_{UU})$
where
\begin{equation}
\ddot{z}_U \triangleq\sum_{k=1}^{K}\dot{z}_{U}^k
\label{eq:gsf} \end{equation}
\begin{equation}
\ddot{\Sigma}_{UU} \triangleq\Sigma_{UU}+\sum_{k=1}^{K} \dot{\Sigma}_{UU}^k \ .
\label{eq:gsc}
\end{equation}
\label{def:gs}
\end{defi}
Using GP-DDF \cite{chen12}, each vehicle exploits the global summary to compute a globally consistent predictive Gaussian distribution of the log-measurements of any set of unobserved regions. The resulting predictive Gaussian mean and variance can then be plugged into (\ref{eq:lgpmu}) to obtain the log-Gaussian posterior mean for predicting the demand of any unobserved region in the original scale.
To improve the predictive power of GP-DDF, 
we develop the following novel GP-DDF$^+$ algorithm that is further augmented by local information.\vspace{2mm}
\begin{defi}[GP-DDF$^+_k$] Given a common support set
$U\subset V$ known to all $K$ vehicles, the global summary
$(\ddot{z}_{U},\ddot{\Sigma}_{UU})$, the local summary
$(\dot{z}_{U}^k,\dot{\Sigma}_{UU}^k)$, a set $D_k\subset V$ of observed regions and a column vector $z_{D_k}$ of corresponding measurements local to vehicle $k$, its 
GP-DDF$^+_k$ algorithm computes a predictive Gaussian distribution
$\mathcal{N}(\overline{\mu}_S^k, \overline{\Sigma}_{SS}^k)$ of the demand measurements of any set $S\subset V$ of unobserved 
regions where $\overline{\mu}^k_S\triangleq\left(\overline{\mu}_s^{k}\right)_{s\in S}$ and $\overline{\Sigma}_{SS}^k \triangleq\left(\overline{\sigma}_{ss'}^k\right)_{s,s'\in S}$ such that
\begin{equation}
%
%
\displaystyle \overline{\mu}_s^k\triangleq \mu_{s} 
+\left({\gamma}_{sU}^{k}\ddot{\Sigma}_{UU}^{-1}\ddot{z}_U
-\Sigma_{sU}\Sigma_{UU}^{-1}\dot{z}_{U}^{k}\right)
+\dot{z}_{s}^{k}
%
\label{eq:lamu}
\end{equation}
\begin{equation}\vspace{-0mm}
%
\begin{array}{c}
\overline{\sigma}_{ss'}^k\triangleq  
\sigma_{ss'}
  -\left(\gamma_{sU}^{k}\Sigma_{UU}^{-1}\Sigma_{Us'}
  -\Sigma_{sU}\Sigma_{UU}^{-1}\dot{\Sigma}_{Us'}^{k}
  \right.
\\
  \hspace{12.3mm}\left.
  -\ {\gamma}_{sU}^{k}\ddot{\Sigma}_{UU}^{-1}{\gamma}_{Us'}^{k}\right)
- \dot{\Sigma}_{ss'}^{k}
%
\end{array}
\label{eq:lacov}
\end{equation}
and
\begin{equation}
\gamma_{sU}^{k}\triangleq
\Sigma_{sU}+\Sigma_{sU}\Sigma_{UU}^{-1}\dot{\Sigma}_{UU}^{k}
-\dot{\Sigma}_{sU}^{k}\ .
\label{eq:keys}
\end{equation}
\label{def:lap}
\end{defi}
%
\emph{Remark} $1$. Both the predictive Gaussian mean $\overline{\mu}_s^k$ (\ref{eq:lamu}) and covariance $\overline{\sigma}_{ss'}^k$(\ref{eq:lacov}) of GP-DDF$^+_k$
exploit summary information (i.e., bracketed term) derived from
global and local summaries and local information (i.e., last
    term) derived from local data.\vspace{1mm}  

\noindent
\emph{Remark} $2$. The predictive Gaussian mean $\overline{\mu}_s^k$ (\ref{eq:lamu}) and variance $\overline{\sigma}_{ss}^k$ (\ref{eq:lacov}) can be plugged into (\ref{eq:lgpmu}) to obtain the log-Gaussian posterior mean for predicting the demand of any unobserved region in the original scale.\vspace{1mm}

\noindent
\emph{Remark} $3$. Since different vehicles exploit different local data, their GP-DDF$^+_k$ algorithms provide inconsistent predictions of the mobility demand pattern.\vspace{1mm}  

It is often desirable to achieve a globally consistent demand prediction among all vehicles. To do this, each unobserved region is simply assigned to the vehicle that predicts its demand best, which can be performed in a decentralized way:\vspace{-2mm}
%
\begin{defi}[Assignment Function]
An assignment function $\tau: V\mapsto \{1\dots K\}$ is defined as
%
\begin{equation}
\tau(s)\triangleq\mathop{\arg\min}_{k\in\{1\dots K\}}
\overline{\sigma}_{ss}^{k}
\label{eq:assign} \end{equation}
for all $s\in S$ where the predictive variance $\overline{\sigma}_{ss}^k$ is defined in (\ref{eq:lacov}). From now on, let $\tau_{s}\triangleq\tau(s)$ for notational simplicity.  
%
%
%
\label{def:assign} \end{defi}
\vspace{2mm}
%
%
%
%
Using the assignment function $\tau$, each vehicle can now compute a globally consistent predictive Gaussian distribution, as detailed in Theorem~\ref{thm:ddf}A below:\vspace{2mm}
\begin{thm}[GP-DDF$^+$]
\label{thm:ddf}
Let a common support set $U\subset V$ and a common assignment function
$\tau$ be known to all $K$ vehicles.\vspace{0mm}
\begin{description}
\item[A.\ \ ] The GP-DDF$^+$ algorithm of each vehicle computes a globally consistent predictive Gaussian distribution $\mathcal{N}(\overline{\mu}_S, \overline{\Sigma}_{SS})$ of the demand measurements of any set $S\subset V$ of unobserved regions where $\overline{\mu}_S\triangleq\left(\overline{\mu}_s^{\tau_s}\right)_{s\in S}$ (\ref{eq:lamu}) and $\overline{\Sigma}_{SS}\triangleq\left(\overline{\sigma}_{ss'}\right)_{s,s'\in S}$ such that\vspace{0mm}
%
%
%
\begin{equation}\vspace{-0mm}
\hspace{0mm}
\overline{\sigma}_{ss'}\triangleq \left\{
\begin{array}{cl}
\overline{\sigma}_{ss'}^{\tau_s}   & \hspace{0mm}\mbox{if\;} \tau_s=\tau_{s'},\vspace{1mm}\\ 
\Sigma_{ss'|U}
+ \gamma^{\tau_s}_{sU}\ddot{\Sigma}_{UU}^{-1}\gamma^{\tau_{s'}}_{Us'}
&\hspace{0mm} \mbox{otherwise\;},
%
\end{array}\right.
\label{eq:dpiccov}
\vspace{-0mm}
\end{equation}
and $\gamma^{\tau_{s'}}_{Us'}$ is the transpose of $\gamma^{\tau_{s'}}_{s'U}$.\vspace{2mm}
\item[B.\ \ ] Let $\mathcal{N}(\mu_{S|D}^{\mbox{\tiny PIC}},
    \Sigma_{SS|D}^{\mbox{\tiny PIC}})$ be the predictive Gaussian
distribution computed by the centralized sparse partially independent
conditional (PIC) approximation of GP model \cite{snelson07} where $\mu_{S|D}^{\mbox{\tiny PIC}}\triangleq\left(\mu_{s|D}^{\mbox{\tiny PIC}}\right)_{s\in S}$ and $\Sigma_{SS|D}^{\mbox{\tiny PIC}}\triangleq\left(\sigma_{ss'|D}^{\mbox{\tiny PIC}}\right)_{s,s'\in S}$ such that
\vspace{-0mm}
\begin{equation}
  \mu_{s|D}^{\mbox{\tiny PIC}} \triangleq
\mu_s+\widetilde{\Gamma}_{sD}\left(
\Gamma_{DD}+\Lambda\right)^{-1}\hspace{-0mm}\left(z_D-\mu_D \right)\vspace{-3mm}
\label{eq:picmu} \end{equation}   
%
\vspace{-0mm}\begin{equation}\vspace{-0mm}
\sigma_{ss'|D}^{\mbox{\tiny PIC}} \triangleq
\sigma_{ss'}-\widetilde{\Gamma}_{sD}\left(\Gamma_{DD}
+\Lambda\right)^{-1}\widetilde{\Gamma}_{Ds'}\vspace{-0mm}
\label{eq:picvar} \end{equation} 
and $\widetilde{\Gamma}_{Ds'}$ is the transpose of $\widetilde{\Gamma}_{s'D}$ such that \vspace{0mm}
\begin{equation}
\Gamma_{BB'}\triangleq \Sigma_{BU}\Sigma_{UU}^{-1}\Sigma_{UB'}\vspace{-0mm} 
\label{eq:sparse} \end{equation} 
\begin{equation}
\widetilde{\Gamma}_{sD}\triangleq (\widetilde{\Gamma}_{s\bar{s}})_{\bar{s}\in D} 
\label{eq:pick} \end{equation} 
\begin{equation}
\widetilde{\Gamma}_{s\bar{s}}\triangleq\bigg{\{}
\begin{array}{cl}
  \sigma_{s\bar{s}}& \mbox{if\;} \tau_{s}=\tau_{\bar{s}},\\
  \Gamma_{s\bar{s}}& \mbox{otherwise\;},
\end{array} \label{eq:pickernel} \end{equation} 
and $\Lambda$ is a block-diagonal matrix constructed from the $K$
diagonal blocks of $\Sigma_{DD|U}$, each of which is a matrix
$\Sigma_{D_k D_k|U}$ for $k=1,\ldots,K$ where $D=\bigcup_{k=1}^K D_k$,
and let $\tau_{\bar{s}}\triangleq k$ for all $\bar{s}\in D_k$.
Then, $\overline{\mu}_s = \mu_{s|D}^{\mbox{\tiny PIC}}$ and $\overline{\sigma}_{ss'} = \sigma_{ss'|D}^{\mbox{\tiny PIC}}$ for all $s, s'\in S$. \vspace{2mm}
\end{description}
\end{thm}
%
%
The proof of Theorem~\ref{thm:ddf}B is given in Appendix A\if\myproof0 of \cite{AA13}\fi.\vspace{1mm}

\noindent
\emph{Remark} $1$. In Theorem~\ref{thm:ddf}A, if 
$\tau_s=\tau_{s'}=k$, then vehicle $k$ can compute $\overline{\mu}^{\tau_s}_s$ (\ref{eq:lamu}) and $\overline{\sigma}_{ss'}$ (\ref{eq:lacov}) locally and send them to the other vehicles that request them. 
Otherwise, $\tau_s\neq\tau_{s'}$ and vehicle $k$ has to request $|U|$-sized vectors
$\gamma_{sU}^{\tau_s}$ and $\gamma_{s'U}^{\tau_{s'}}$ from the respective vehicles $\tau_s$ and $\tau_{s'}$ to compute $\overline{\sigma}_{ss'}$ (\ref{eq:dpiccov}).\vspace{1mm}

\noindent
\emph{Remark} $2$. The equivalence result of Theorem \ref{thm:ddf}B implies that the computational load of the centralized PIC approximation of GP can be distributed among $K$ vehicles, hence improving the time efficiency of demand prediction.
Supposing $|S|\leq |U|$ and $|S|\leq|D|/K$ for simplicity, 
the 
 $\mathcal{O}\hspace{-1mm}\left( |D|((|D|/K)^2 + |U|^2) \right)$
time incurred by PIC can be reduced to $\mathcal{O}\hspace{-1mm}\left( (|D|/K)^3 + |U|^3 +|U|^2K \right)$
time of running GP-DDF$^+$ on each of the $K$ vehicles.  Hence, GP-DDF$^+$ scales better with increasing size $|D|$ of data.\vspace{1mm}

\noindent \emph{Remark} $3$. The equivalence result also  sheds some light on an important property of GP-DDF$^+$ based on the structure of PIC: 
It is assumed that
$Z_{D_1\bigcup S_1}$, \ldots, $Z_{D_K\bigcup S_K}$ are conditionally independent given the support set $U$. As compared to GP-DDF that assumes conditional
independence of $Z_{D_1},\ldots, Z_{D_K}, Z_{S_1}, \ldots, Z_{S_K}$, GP-DDF$^+$ can predict $Z_{S}$ better since it imposes a weaker conditional independence assumption. 
Experimental results on real-world mobility demand data (Section~\ref{sec:exp}) also show that GP-DDF$^+$ achieves predictive accuracy comparable to FGP and significantly better than GP-DDF, thus justifying the practicality of such an assumption for predicting a mobility demand pattern. 
%
\section{Decentralized Active Demand Sensing} \label{sec:das}
%
%
%
Suppose that there are $K$ vacant MoD vehicles in the fleet actively cruising the service area and 
%
each vehicle $k\in\{1\dots K\}$ has observed a set $D_k \subset V$ of regions.
In the active demand sensing problem, all vehicles have to jointly select the most informative walks $w_1^*,\dots w_K^*$ of length $H$ each along which demand data will be sampled:
\begin{equation}
\displaystyle
 (w_1^*,\dots,w_K^*)\triangleq\mathop{\arg\max}_{(w_1,\dots,w_K)}
 \mathbb{H}\left[Y_{\bigcup_{k=1}^K S_{w_k}}\Big|Y_{\bigcup_{k=1}^K
   {D_k}}\right]
\label{eq:maxent} 
\end{equation}
where $S_{w_k}$ denotes the set of unobserved regions to be visited by the walk $w_k$.  To ease notation, let $w\triangleq(w_1\dots w_K)$ and $S_w=\bigcup_{k=1}^K S_{w_k}$ (similarly, for $w^*$ and $S_{w^*}$). 
Then, each vehicle $k$ executes its walk $w^*_k$ while observing the
demands of regions $S_{w^*_k}$, and updates its location and stored data.

To derive the most informative joint walk $w^*$, the posterior entropy (\ref{eq:maxent}) of every possible joint walk $w$ has to be evaluated.
Such a centralized strategy cannot be performed in real time due to the following two issues: (a) It relies on all the demand data that are gathered in a distributed manner by the
vehicles, thus incurring huge time and communication overheads with large data, and (b) it involves evaluating a prohibitively large number of joint walks (i.e.,
exponential in the fleet size).  

The first issue can be alleviated by approximating the log-Gaussian posterior joint
entropy using the decentralized GP-DDF$^+$ algorithm (Theorem~\ref{thm:ddf}A), thus distributing its computational load among all vehicles.  
Then, the active demand sensing problem (\ref{eq:maxent}) is approximated by 
%
\begin{equation}\vspace{-0mm}
w^*=\mathop{\arg\max}_w \overline{\mathbb{H}}\left[Y_{S_w}\right] 
\label{eq:approxmaxent} 
\end{equation}
\begin{equation}\vspace{-0mm}
\overline{\mathbb{H}}\left[Y_{S_w}\right]\triangleq
\frac{1}{2}\log\hspace{-0.5mm}\left(2\pi e\right)^{\left|S_{w}\right|}
\left|\overline{\Sigma}_{S_wS_w}\right| + \overline{\mu}_{S_w} \cdot
\mathbf{1} \;. 
\label{eq:approxent} \end{equation}
%
To obtain $\overline{\mathbb{H}}\left[Y_{S_w}\right]$ (\ref{eq:approxent}),  
$\Sigma_{S_w S_w|D}$ and $\mu_{S_w|D}$ in $\mathbb{H}[Y_{S_w}|Y_D]$ ((\ref{eq:lgpent}) \& (\ref{eq:maxent})) are replaced by $\overline{\Sigma}_{S_wS_w}$ and
$\overline{\mu}_{S_w}$ defined in Theorem~\ref{thm:ddf}A, respectively.

%
To address the second issue, a partially decentralized
active sensing strategy proposed by 
\cite{chen12}
partitions the vehicles into several
small groups such that each group of vehicles selects its joint walk
independently. This partitioning heuristic performs poorly when
the largest group formed still contains many vehicles. In our work, this is indeed the case because many vehicles tend to cluster within hotspots, as explained later.
To scale well in the fleet size, we therefore adopt a fully decentralized active sensing (DAS) strategy by assuming that the joint walk $w_1^*\dots w_K^*$ is derived by selecting the locally optimal walk of each vehicle $k$:
\begin{equation}\vspace{-0mm}
w_k^* = \mathop{\arg\max}_{w_k}
\overline{\mathbb{H}}\left[Y_{S_{w_k}}\right]
\label{eq:dmaxent} 
\end{equation}
where $\overline{\mathbb{H}}\left[Y_{S_{w_k}}\right]$ is defined in the same way as (\ref{eq:approxent}).
Then, each vehicle can select its locally optimal walk independently of
the other vehicles, thus significantly reducing the search space of joint
walks. A consequence of such an assumption is that,  without coordinating their walks, the vehicles may select suboptimal joint walks (e.g., two vehicles' locally optimal walks are highly correlated).  In practice, this assumption becomes less restrictive when the size $|D|$ of data increases to potentially reduce the degree of violation of conditional independence of $Y_{S_{w_1}},\ldots,Y_{S_{w_K}}$.

More importantly, it can be observed from (\ref{eq:approxent}) and
(\ref{eq:dmaxent}) that the cruising behavior of DAS trades off
between exploring sparsely sampled regions with high predictive uncertainty (i.e., by maximizing the log-determinant of Gaussian posterior covariance matrix $\overline{\Sigma}_{S_{w_k}S_{w_k}}$ term) and hotspots (i.e., by maximizing the Gaussian posterior mean vector $\overline{\mu}_{S_{w_k}}$ term). As a result, it redistributes vacant MoD vehicles to regions with high likelihood of picking up users.
Hence, besides gathering the most informative data for predicting the mobility demand pattern, DAS is able to achieve a dual effect of fleet rebalancing to service mobility demands. 
%
%
%
%
\section{Time and Communication Analysis}
\label{sec:overhead}
%
%
In this section, we analyze the time and communication overheads of our proposed
GP-DDF$^+$ coupled with DAS algorithm (Algo.~\ref{algo:d2fp}) and
compare them with that of both FGP (Section~\ref{sec:lgp}) and GP-DDF
\cite{chen12} coupled with DAS algorithms (Section~\ref{sec:das}).
%
\begin{algorithm} {\scriptsize \DontPrintSemicolon
\While{true} {
\tcc{Data fusion (Section~\ref{sec:ddf})}
Construct local summary by (\ref{eq:lsf}) \& (\ref{eq:lsc})\;    
Exchange local summary with every vehicle $i\neq k$\; 
Construct global summary by (\ref{eq:gsf}) \& (\ref{eq:gsc})\; 
Construct assignment function by (\ref{eq:assign}) \; 
Predict demand measurements of unobserved regions by (\ref{eq:lamu}) \& (\ref{eq:dpiccov})\;
\tcc{Active Sensing (Section~\ref{sec:das})}
Compute local maximum-entropy walk $w^{*}_k$ by (\ref{eq:dmaxent})\;
Execute walk $w^*_k$ and observe its demand measurements $Y_{w_k^*}$\;
Update local information $D_k$ and $y_{D_k}$ \; 

}
\caption{GP-DDF$^+$+DAS$(U, K, H, k, D_k, z_{D_k})$}
}
\label{algo:d2fp} \end{algorithm}
\subsection{Time Complexity} 
Firstly, each vehicle $k$ has to evaluate $\Sigma_{D_kD_k|U}$ in
$\mathcal{O}\hspace{-1mm}\left( |U|^3+|U|(|D|/K)^2\right)$ time and invert it in
$\mathcal{O}\hspace{-1mm}\left((|D|/K)^3\right)$ time. 
After that, the data fusion component constructs the local summary in
$\mathcal{O}\hspace{-1mm}\left(|U|^2|D|/K+|U|(|D|/K)^2 \right)$ time by
(\ref{eq:lsf}) and (\ref{eq:lsc}), and subsequently the global summary
in $\mathcal{O}\hspace{-1mm}\left( |U|^2K \right)$ time by (\ref{eq:gsf}) and
(\ref{eq:gsc}).
To construct the assignment function for any
unobserved set $S\subset V$, vehicle $k$ first computes $|S|$ number of
$\gamma_{sU}^k$ for all unobserved regions $s\in S$ in
$\mathcal{O}\hspace{-1mm}\left(|S||U|^2+|S|(|D|/K )^2\right)$ time by
(\ref{eq:keys}). Then, after inverting $\ddot{\Sigma}_{UU}$ in
$\mathcal{O}(|U|^3)$, the predictive means and variances for all $s\in S$ are computed
in $\mathcal{O}\hspace{-1mm}\left(|S||U|^2 +|S|(|D|/K)^2 \right)$ time by
(\ref{eq:lamu}) and (\ref{eq:dpiccov}), respectively.
Let $\Delta\triangleq \delta^H$ denote the number of possible walks of
length $H$ where $\delta$ is the maximum out-degree of graph $G$. In the
active sensing component, 
to obtain the locally optimal walk, the log-Gaussian posterior entropies
(\ref{eq:dmaxent}) of all possible walks are derived from
(\ref{eq:lamu}) and (\ref{eq:dpiccov}), respectively, in
$\mathcal{O}\hspace{-1mm}\left( \Delta H |U|^2 \right)$ and $\mathcal{O}\hspace{-1mm}\left(
\Delta( H |U| )^2\right)$ time.  We assume $|S| \leq
\delta\Delta$ where $S$ denotes the set $\bigcup_{w_k} S_{w_k}$ of regions
covered by any vehicle $k$'s all possible walks of length $H$. Then, the
time complexity for our GP-DDF$^+$ coupled with DAS algorithm is 
$\mathcal{O}\hspace{-1mm}\left( ( |D|/K )^3\hspace{-0.7mm}
+\hspace{-0.6mm}|U|^3\hspace{-0.7mm}+\hspace{-0.6mm}|U|^2K\hspace{-0.6mm} +\hspace{-0.6mm}\Delta( H^3\hspace{-0.7mm} +\hspace{-0.6mm} (H|U|)^2 \hspace{-0.7mm}+\hspace{-0.7mm}(|D|/K )^2) \right)$.  
In contrast, the time incurred by FGP and GP-DDF
coupled with DAS algorithms are, respectively, $\mathcal{O}\hspace{-1mm}\left(
|D|^3 + \Delta (H^3+ (H|D|)^2)\right)$ and $\mathcal{O}\hspace{-1mm}\left( ( |D|/K
)^3 + |U|^3 +|U|^2K + \Delta (H^3+ (H|U|)^2) \right)$.
It can be observed that our GP-DDF$^+$ coupled with DAS algorithm can scale better with large size $|D|$ of data and fleet size $K$ than FGP coupled with DAS algorithm, and its increased computational load, as compared to GP-DDF coupled with DAS algorithm, is well distributed among $K$ vehicles. 
%
\subsection{Communication complexity} 
In each iteration, each vehicle of the system running our GP-DDF$^+$
coupled with DAS algorithm has to broadcast a $\mathcal{O}(|U|^2)$-sized
local summary for constructing the global summary, exchange
$\mathcal{O}(\Delta)$ scalar values for constructing the assignment function, and request $\mathcal{O}(\Delta)$ number
of $\mathcal{O}(|U|)$-sized $\gamma_{sU}^k$ components for evaluating
the entropies of all possible local walks. In contrast, FGP coupled with DAS algorithm needs to broadcast $\mathcal{O}(|D|/K)$-sized message
comprising all its local data to handle communication failure, and GP-DDF
coupled with DAS algorithm only needs to broadcast a $\mathcal{O}(|U|^2)$-sized local
summary. 

\section{Experiments and Discussion} \label{sec:exp} 
\begin{figure}[h] \begin{tabular}{cc}
%
\hspace{-8mm} \includegraphics[scale=0.32]{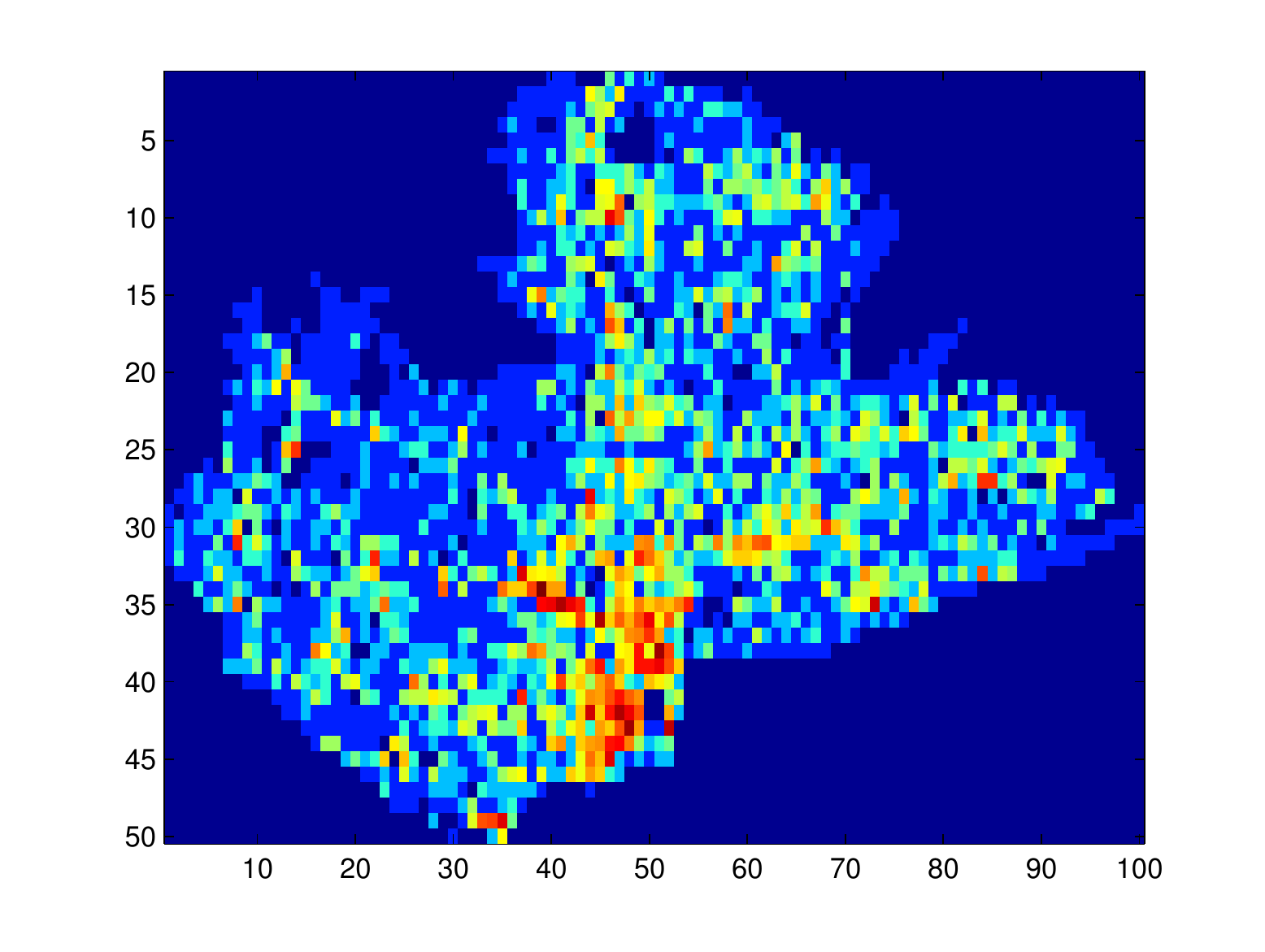} &
\hspace{-9.5mm} \includegraphics[scale=0.32]{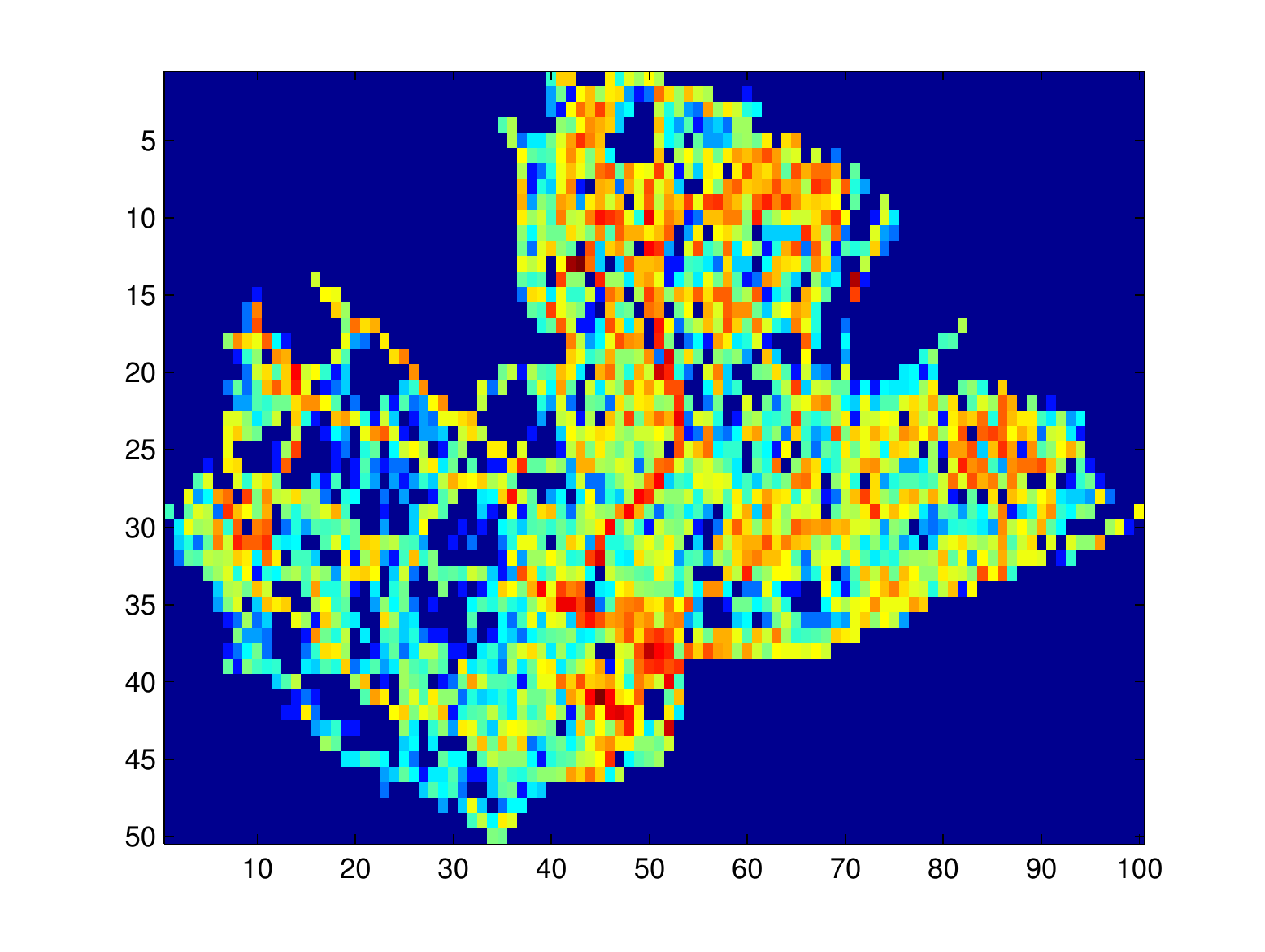}\vspace{-3mm}\\ 
%
\hspace{-8mm} (a) Demand & \hspace{-9.5mm} (b) Supply
\end{tabular}
\caption{Historic demand and supply distributions.}
\label{fig:dmd} \end{figure}

%
This section evaluates the performance of our proposed algorithm using a
real-world taxi trajectory dataset taken from the central business
district of Singapore between $9$:$30$~p.m. and $10$~p.m. on August $2$, $2010$.
The service area is gridded into $50\times 100$ regions such that $2506$
regions are included into the dataset as the remaining regions contain
no road segment for cruising vehicles to access. The maximum out-degree $\delta$
of graph $G$ over these regions is $8$. The feature vector of each region is
specified by its corresponding location.
In any region, the demand (supply) measurement is obtained by counting
the number of pickups (taxis cruising by) from all historic
taxi trajectories generated by a major taxi company in a $30$-minute
time slot. After processing the taxi trajectories, the
historic demand and supply distributions are obtained, as shown in Fig.~\ref{fig:dmd}.
%
Then, a number $C$ of users are randomly distributed over the service area
with their locations drawn from the demand distribution (Fig.
\ref{fig:dmd}a). Similarly, a fleet of $K$ vacant MoD vehicles are
initialized at locations drawn from the supply distribution
(Fig. \ref{fig:dmd}b). 

In our simulation, when a vehicle enters a region with users, it
picks up one of them randomly. Then, the MoD system removes this vehicle
from the fleet of vacant cruising vehicles and introduce a new vacant
vehicle drawn from the supply distribution. Similarly, a new user
appears at a random location drawn from the demand distribution.
The MoD system operates for $L$ time steps and each vehicle plans a walk
of length $4$ at each time step, with all vehicles running a data fusion
algorithm coupled with our DAS strategy. We will
compare the performance of our GP-DDF$^+$ algorithm with that of FGP
and GP-DDF algorithms when coupled with our DAS strategy.
The experiments are conducted on a Linux system with Intel$\circledR$
Xeon$\circledR$ CPU E5520 at 2.27 GHz.
%
\subsection{Performance Metrics} 
The tested algorithms are evaluated with two sets of performance metrics.  The
performance of sensing and predicting mobility demands is
evaluated using (a) root mean square error (RMSE)
$\sqrt{|V|^{-1}\sum_{s\in V}\left(y_s-\widehat{\mu}_{s|D}\right)^2}$
where $y_s$ is the demand measurement and $D$ is the set of regions observed by the MoD vehicles, and (b) incurred time of the algorithms.

The performance of servicing mobility demands is evaluated by comparing
the  Kullback-Leibler divergence (KLD) $\sum_{s\in
V}P_c(s)\log\left(P_c(s)/P_d(s)\right)$ between the fleet distribution
$P_c$ of vacant MoD vehicles controlled by the tested algorithms and historic demand
distribution $P_d$ (i.e., lower KLD implies better balance between fleet and demand), average cruising length of MoD vehicles, average waiting time of
users, and total number of pickups resulting from the tested
algorithms.
%
%
\subsection{Results and Analysis} 
%
For notational simplicity, we will use GP-DDF$^+$, FGP, and GP-DDF to
represent the algorithms of their corresponding data fusion components coupled
with DAS strategy in this subsection.

\subsubsection{Performance}
The MoD system comprises $K=20$ vehicles running three
tested algorithms for $L=960$ time steps in a service area with $C=200$ 
users. All results are obtained by averaging over $40$ random instances.

\begin{figure*}[ht] \begin{tabular}{cccccc}
\hspace{-3.5mm} 
    \includegraphics[scale=0.157]{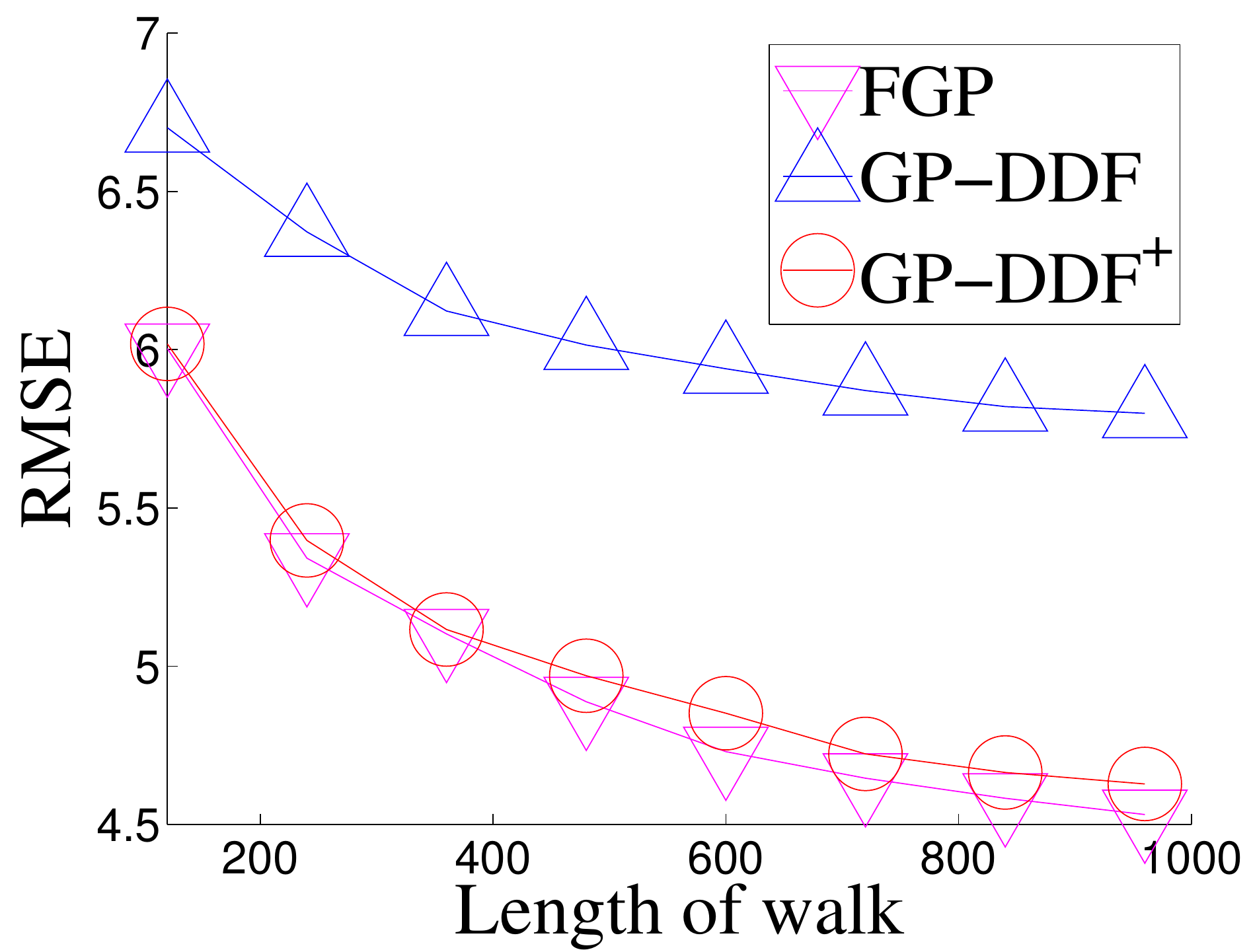}&
\hspace{-5mm} 
    \includegraphics[scale=0.157]{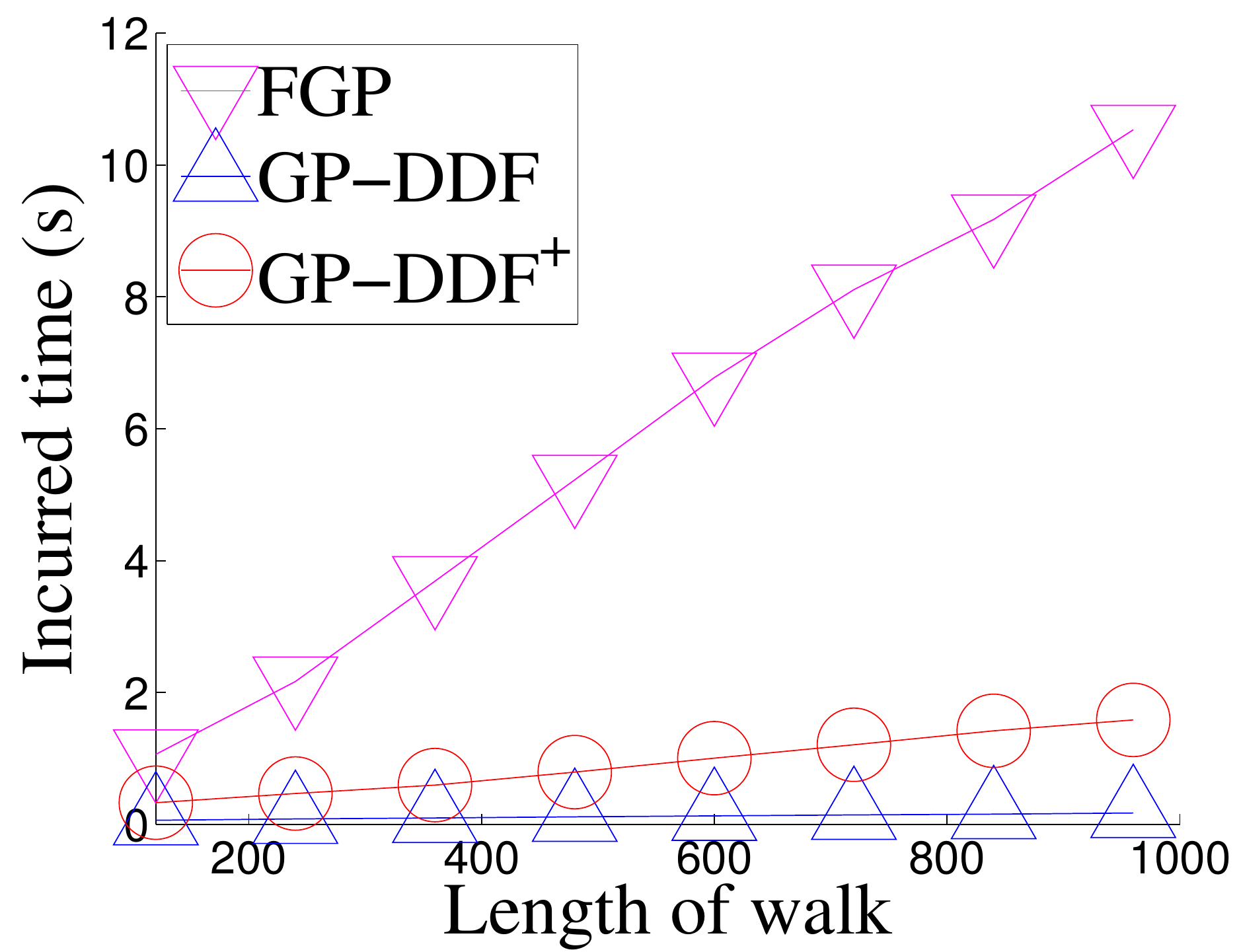}&
\hspace{-5mm} 
    \includegraphics[scale=0.157]{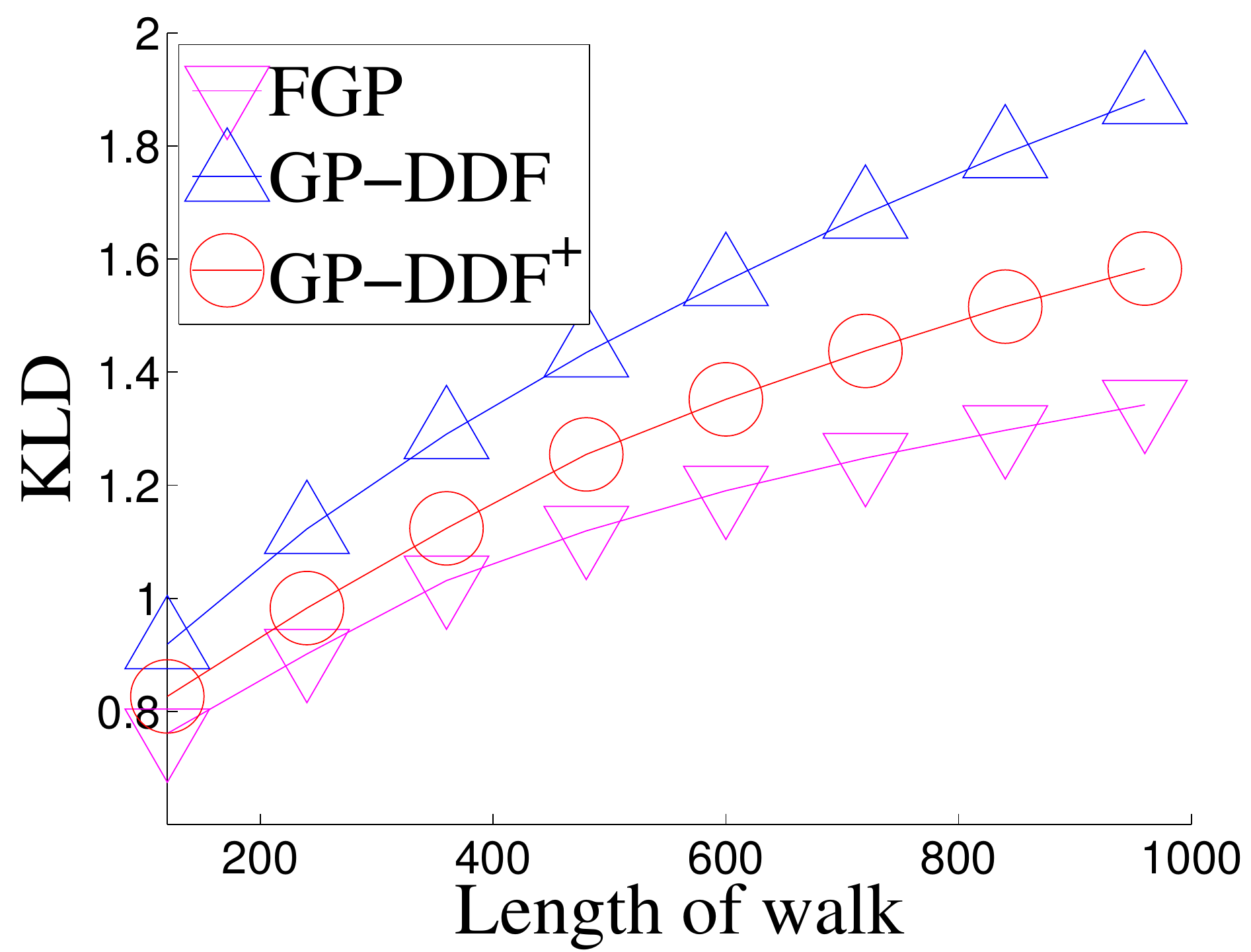}&
\hspace{-5mm} 
    \includegraphics[scale=0.157]{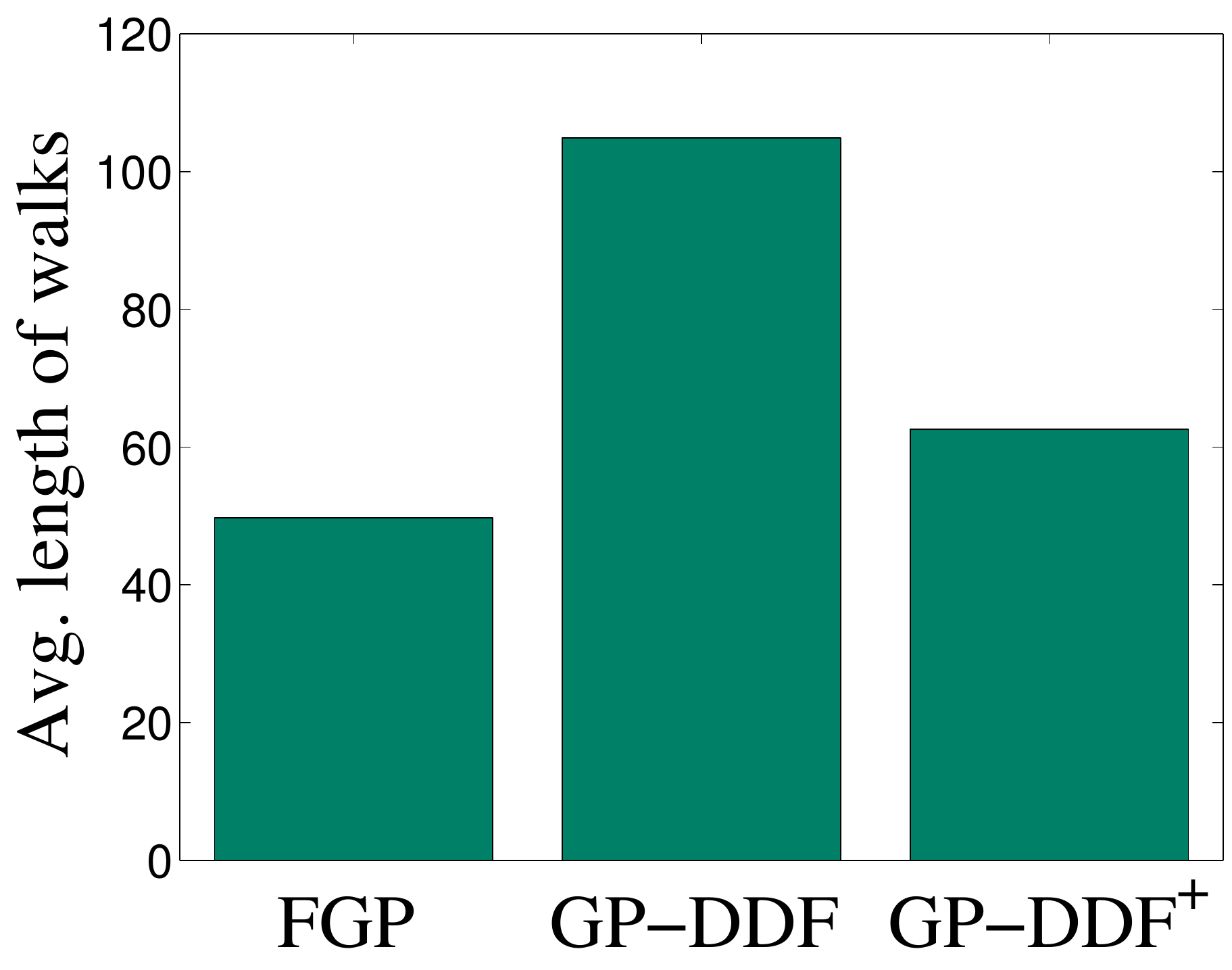}&
\hspace{-4.05mm} 
    \includegraphics[scale=0.157]{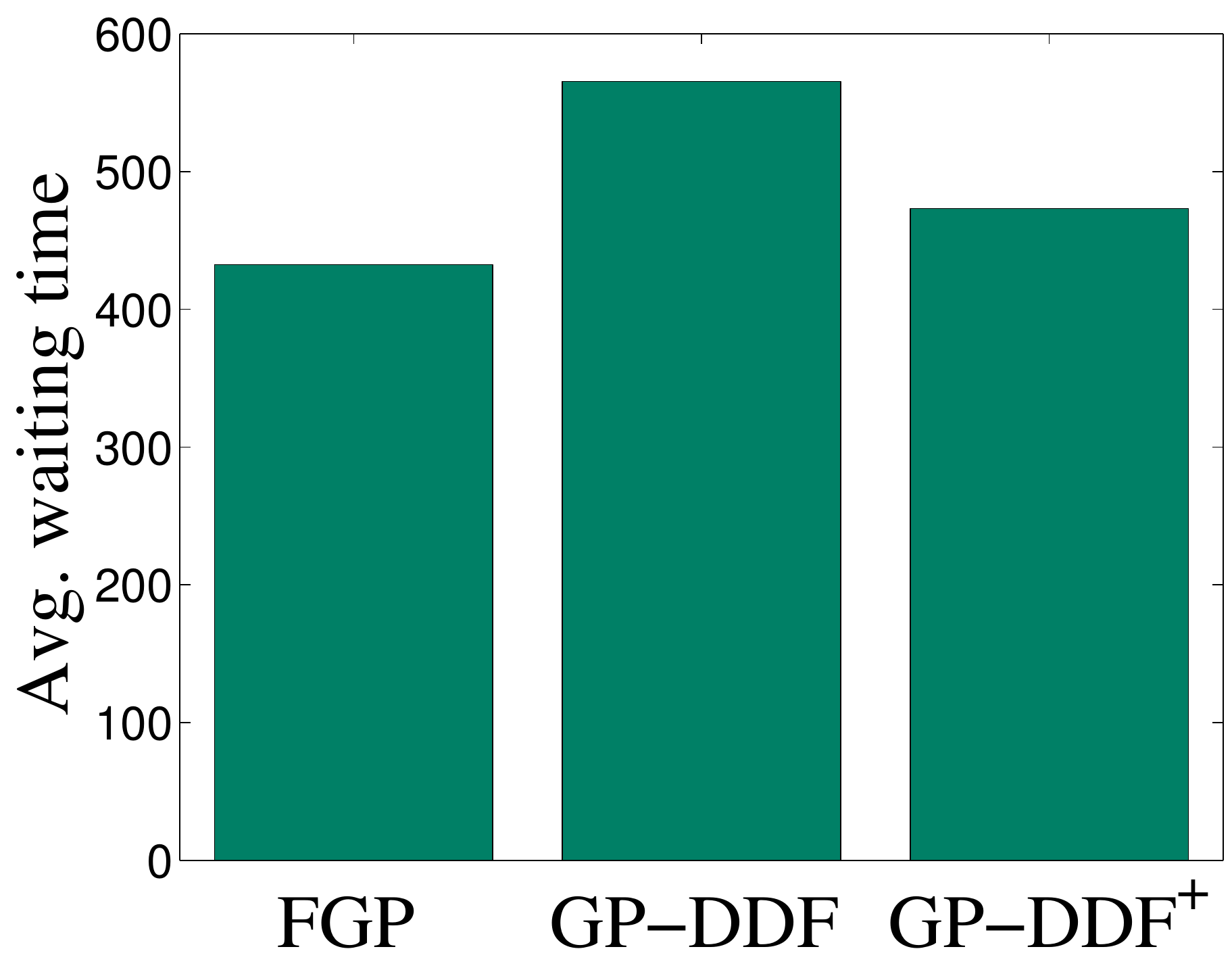}&
\hspace{-4.05mm} 
    \includegraphics[scale=0.157]{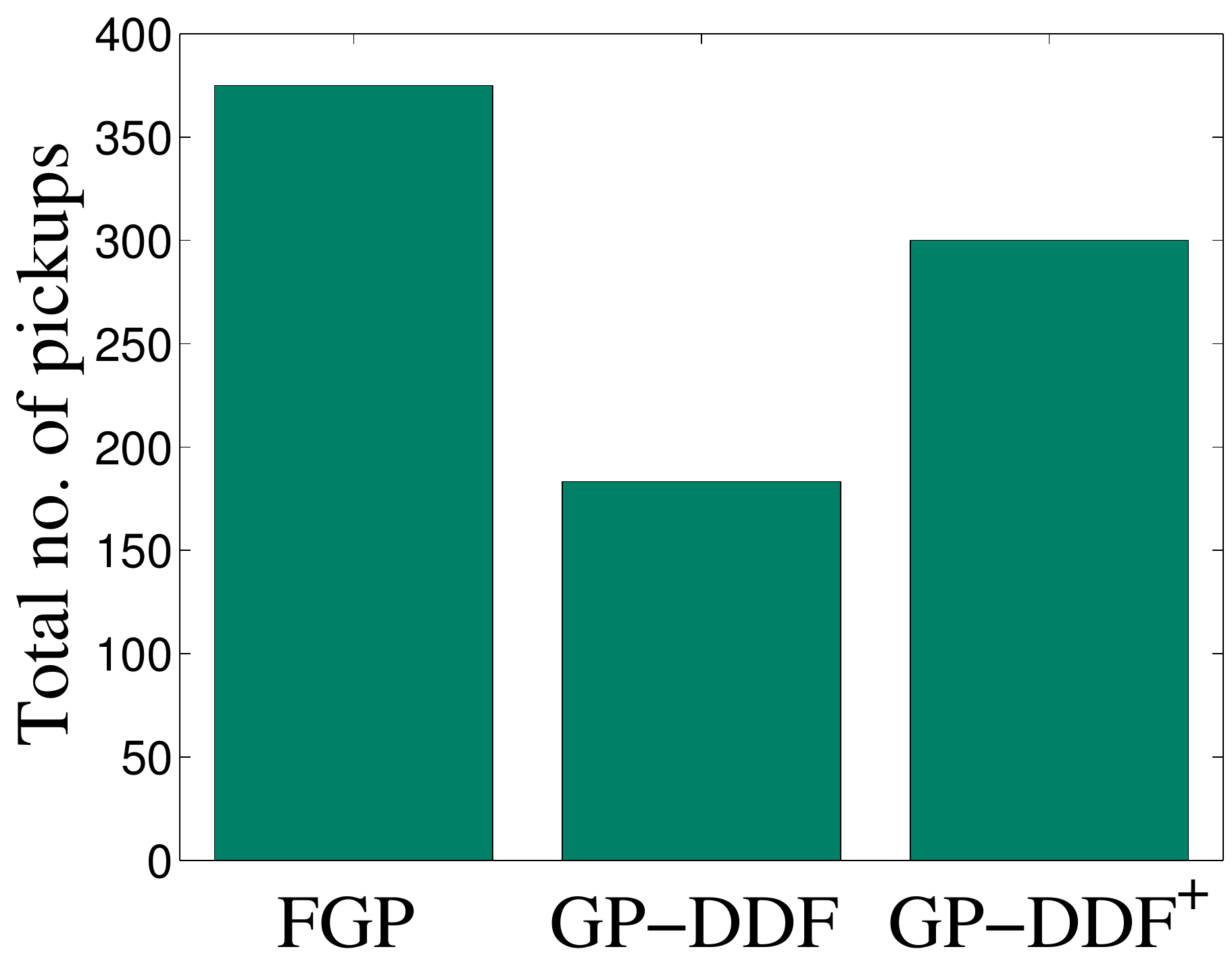}\\
\hspace{-3.5mm} 
    (a) Accuracy & 
\hspace{-5mm} 
    (b) Efficiency &
\hspace{-5mm} 
    (c) Balance & 
\hspace{-5mm} 
    (d) Vehicles &
\hspace{-4.05mm} 
    (e) Users & 
\hspace{-4.05mm} 
    (f) Pickups \\
\end{tabular}
\caption{Performance of MoD systems in sensing, predicting, and servicing mobility demands.}
\vspace{-5mm}

\label{fig:rst1} \end{figure*}

The performance of MoD systems in sensing and predicting mobility
demands is illustrated in Figs. \ref{fig:rst1}a-\ref{fig:rst1}b.
Fig.~\ref{fig:rst1}a shows that the demand data collected by MoD
vehicles using GP-DDF$^+$ can achieve predictive accuracy comparable to
that of using FGP and significantly better than that of using GP-DDF.
This indicates that exploiting the local data of vehicles for predicting
demands of nearby unobserved regions can improve the prediction of the
mobility demand pattern.
Fig.~\ref{fig:rst1}b shows the average incurred time of each vehicle
using three algorithms. GP-DDF$^+$ is significantly more time-efficient
(i.e., one order of magnitude) than FGP, and only slightly less
time-efficient than GP-DDF.  This can be explained by the time analysis
in Section~\ref{sec:overhead}.  
The above results indicate that GP-DDF$^+$ is more practical for
real-world deployment due to a better balance between predictive
accuracy and time efficiency.

The performance of MoD systems in servicing the mobility demands is
illustrated in Figs.~\ref{fig:rst1}c-\ref{fig:rst1}f.
Fig.~\ref{fig:rst1}c shows that a MoD system using GP-DDF$^+$ can
achieve better fleet rebalancing of vehicles to service mobility demands
than GP-DDF, but worse rebalancing than FGP. This implies that a better
prediction of the underlying mobility demand pattern
(Fig.~\ref{fig:rst1}a) can lead to better fleet rebalancing.
Note that KLD (i.e., imbalance between mobility demand and fleet)
  increases over time because we assume that when a vehicle picks up a
  user, its local data is removed from the fleet of cruising vehicles,
  and a new vehicle is introduced at a random location that may be
  distant from a demand hotspot, hence worsening the imbalance between
  demand and fleet.
It can also be observed that an algorithm generating a better balance between fleet
and demand will also perform better in servicing the mobility
demands, that is, shorter average cruising trajectories of vehicles
(Fig.~\ref{fig:rst1}d), shorter average waiting time of users
(Fig.~\ref{fig:rst1}e), and larger total number of pickups
(Fig.~\ref{fig:rst1}f).  These observations imply that exploiting an
active sensing strategy to collect the most informative demand data for
predicting the mobility pattern achieves a dual effect of improving
performance in servicing the mobility demands since these vehicles
have higher chance of picking up users in demand hotspots or
sparsely sampled regions (Section~\ref{sec:das}).
%
%
%
%

\subsubsection{Scalability}
We vary the number $K=10,20,30$ of vehicles in the MoD system, and keep the
total length of walks of all the vehicles to be the same, that is, these
vehicles will walk for $L=960,480,320$ steps, respectively. All three algorithms are
tested in a service area with $C=600$ users. 
All results are obtained by averaging over $40$ random instances.

\begin{figure}[ht] \begin{tabular}{ccc}
\hspace{-3.5mm} 
    \includegraphics[scale=0.15]{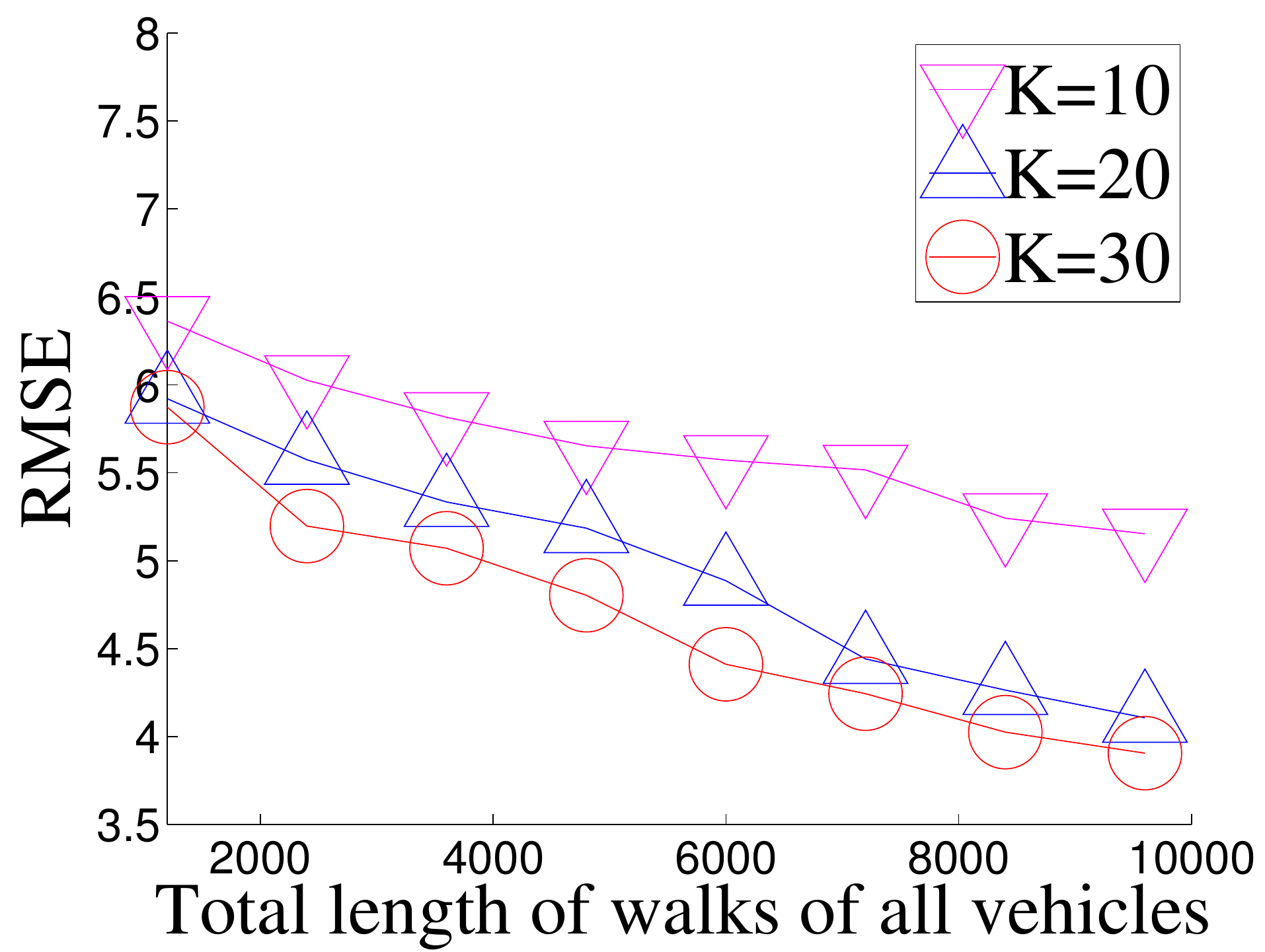}&
\hspace{-5mm} 
    \includegraphics[scale=0.15]{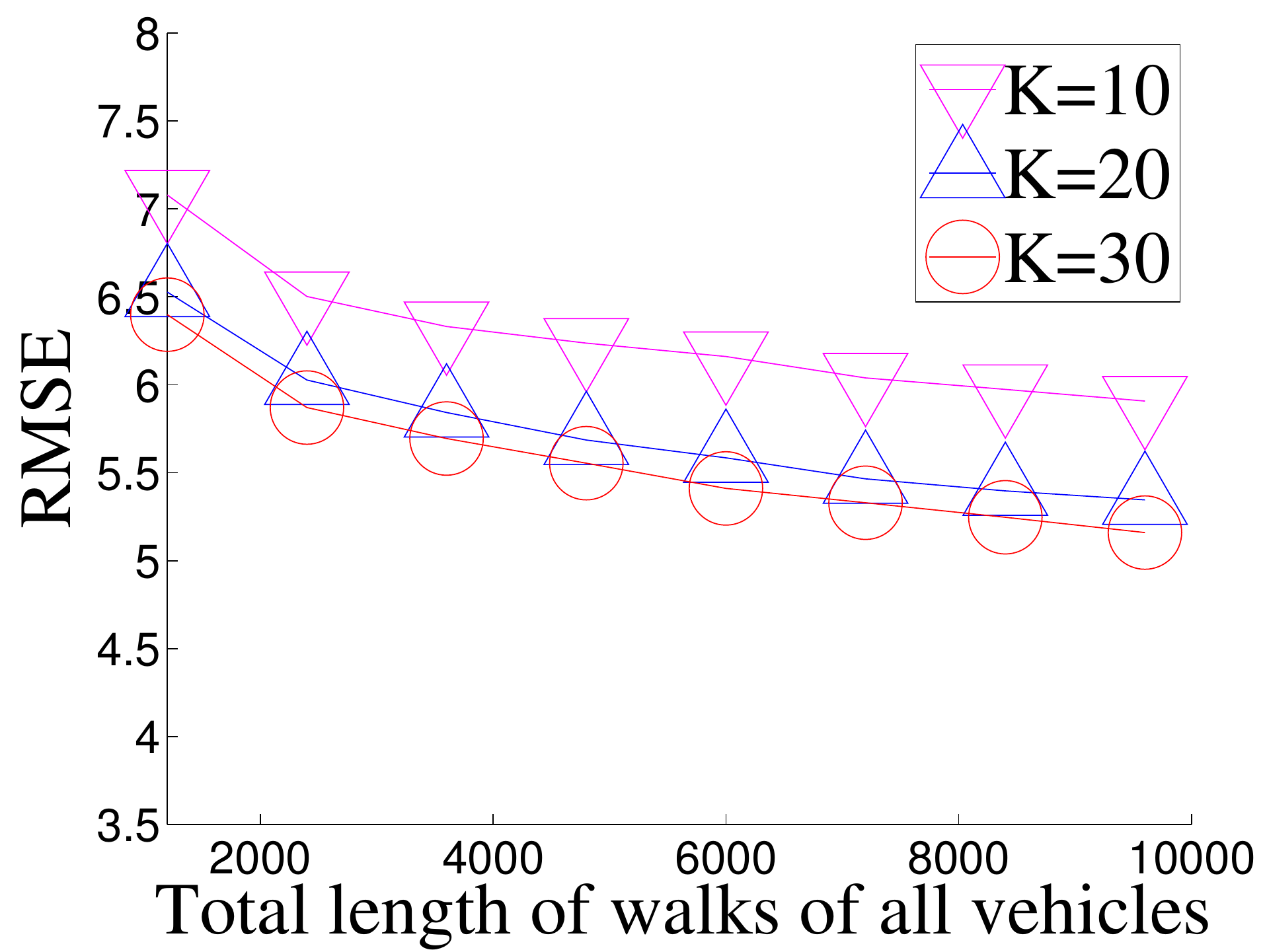}&
\hspace{-5mm} 
    \includegraphics[scale=0.15]{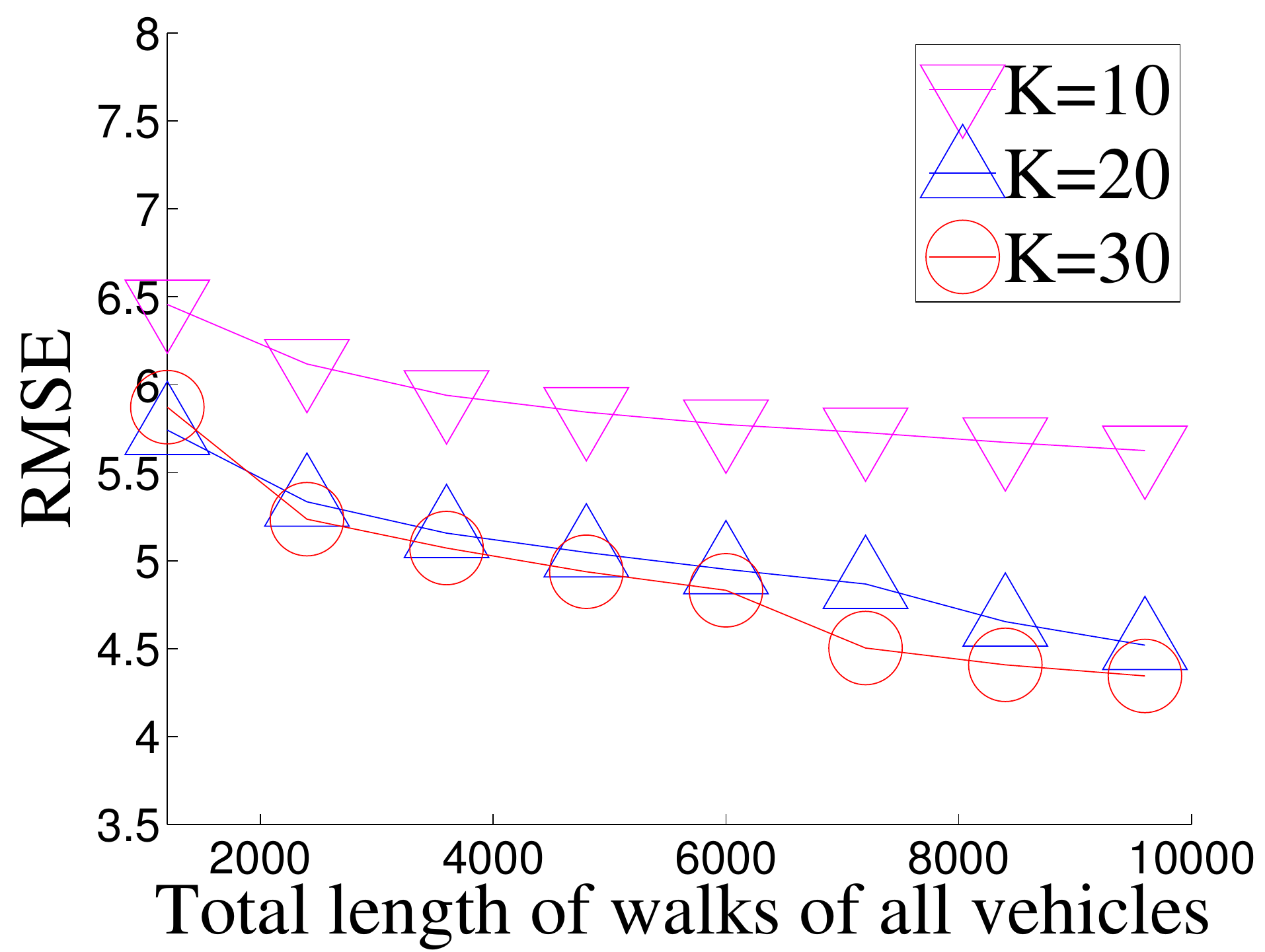}\\
    \hspace{-3.5mm}(a) FGP&
    \hspace{-5mm}(b) GP-DDF&
    \hspace{-5mm}(c) GP-DDF$^+$\\
\hspace{-3.5mm} 
    \includegraphics[scale=0.15]{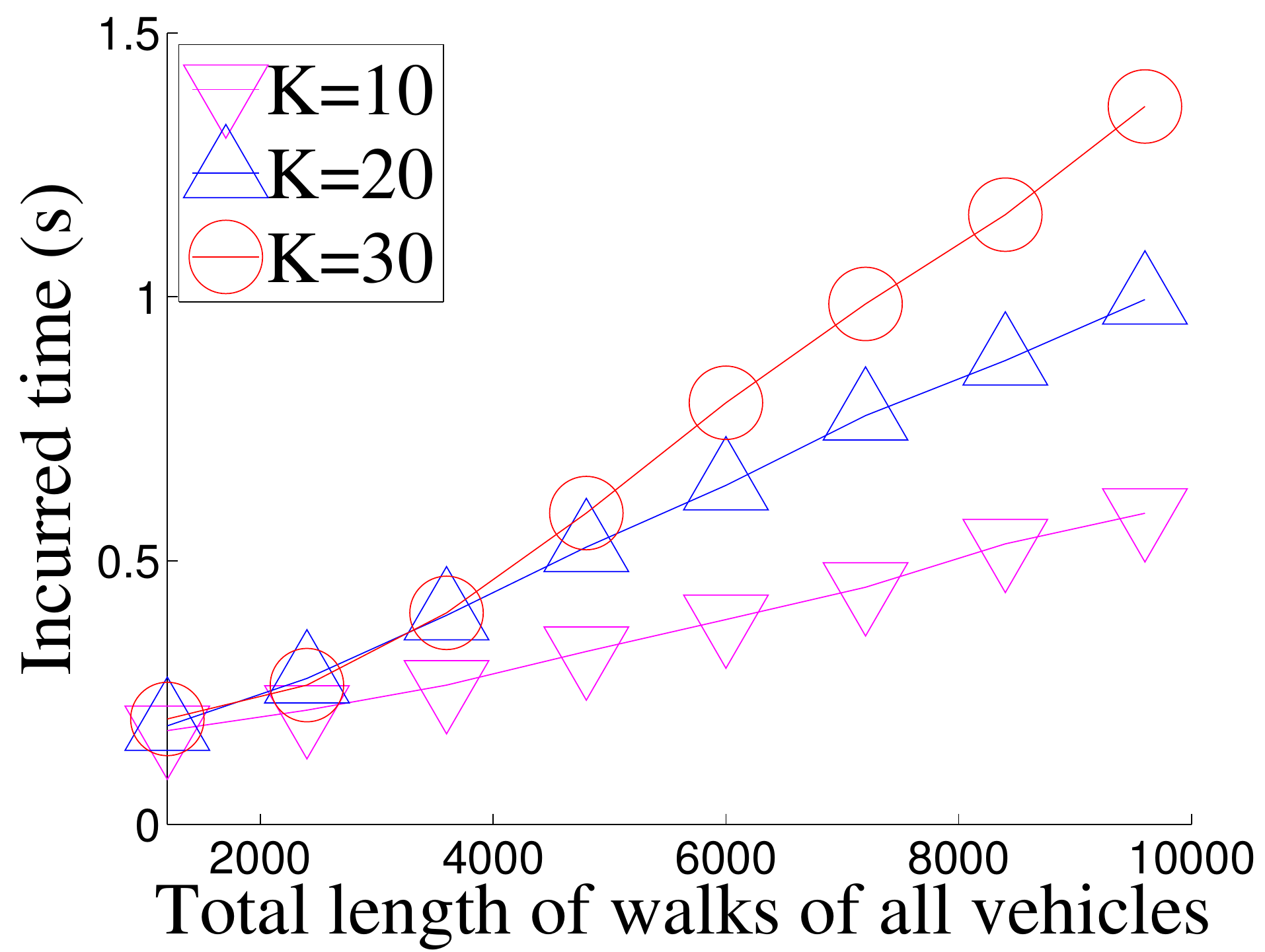}&
\hspace{-5mm} 
    \includegraphics[scale=0.15]{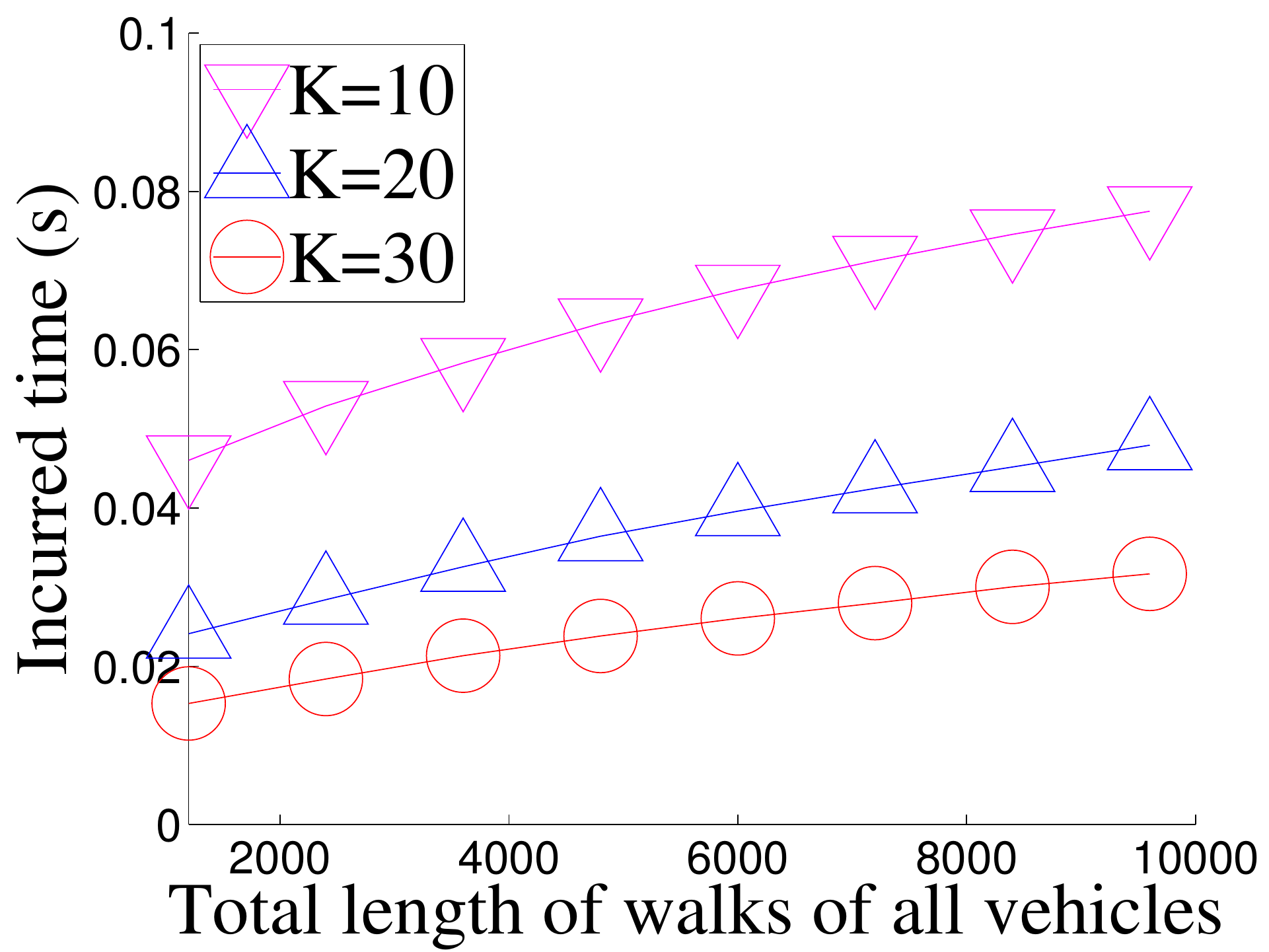}&
\hspace{-5mm} 
    \includegraphics[scale=0.15]{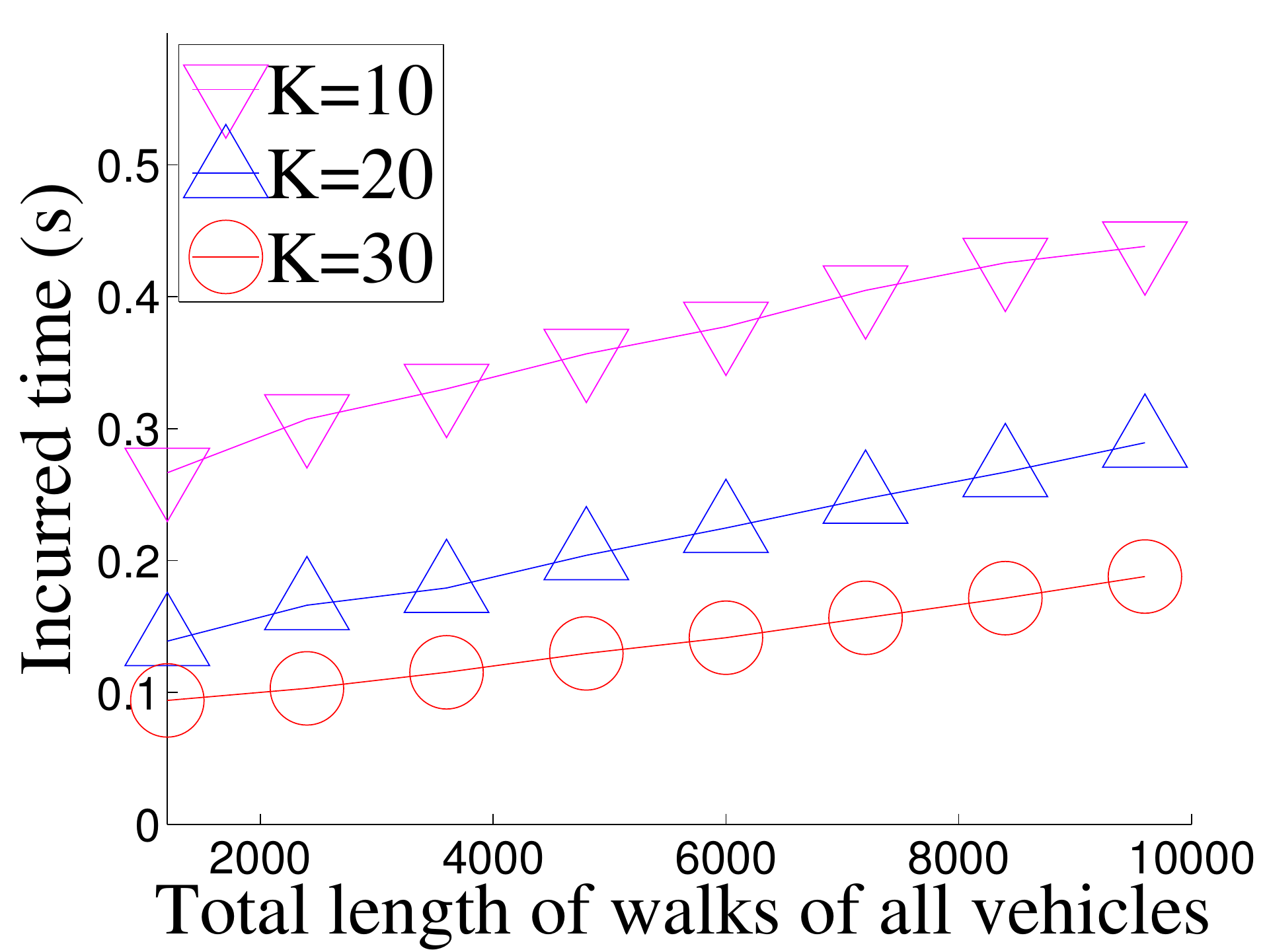}\\
    \hspace{-3.5mm}(d) FGP&
    \hspace{-5mm}(e) GP-DDF&
    \hspace{-5mm}(f) GP-DDF$^+$
%
\end{tabular}
\caption{Scalability of MoD systems in sensing and predicting mobility demands.}
\label{fig:rst3} \end{figure}

From Figs.~\ref{fig:rst3}a-\ref{fig:rst3}c, it can be observed that all three
algorithms can improve their prediction accuracy with an increasing number
of vehicles in the MoD system because more vehicles indicate less walks when the total length of walks are the same, thus suffering less from the myopic planning ($H=4$) and gathering more informative demand data.
Figs.~\ref{fig:rst3}d-\ref{fig:rst3}f show that, with more MoD vehicles, GP-DDF$^+$ and GP-DDF  incur less time, while FGP incurs more time.  This is because the computational load in decentralized data fusion algorithms are distributed among all vehicles,
thus reducing the incurred time with more vehicles.

\begin{figure}[h] \begin{tabular}{ccc}
\hspace{-3.5mm} 
\includegraphics[scale=0.15]{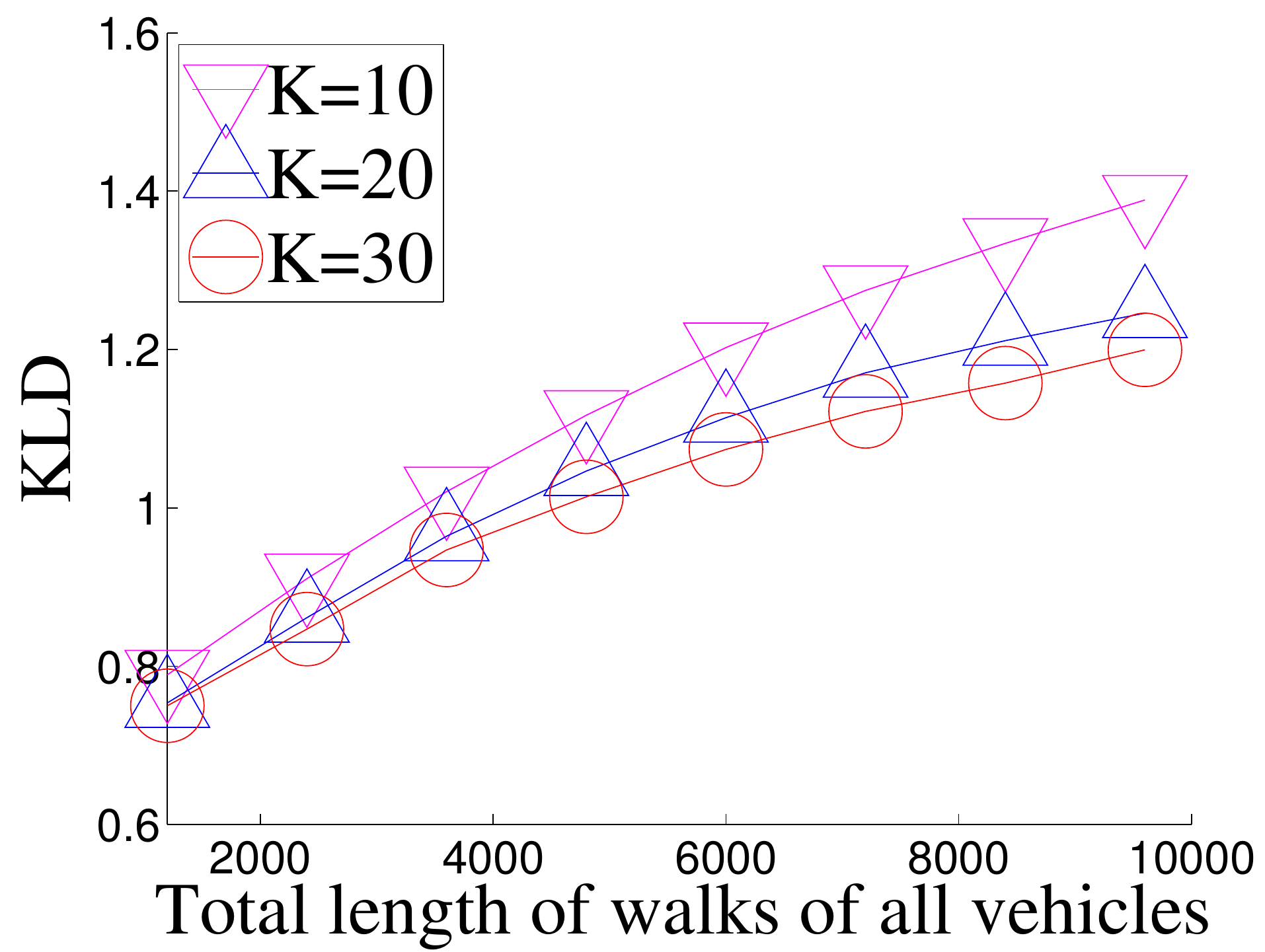}&
\hspace{-5mm} 
\includegraphics[scale=0.15]{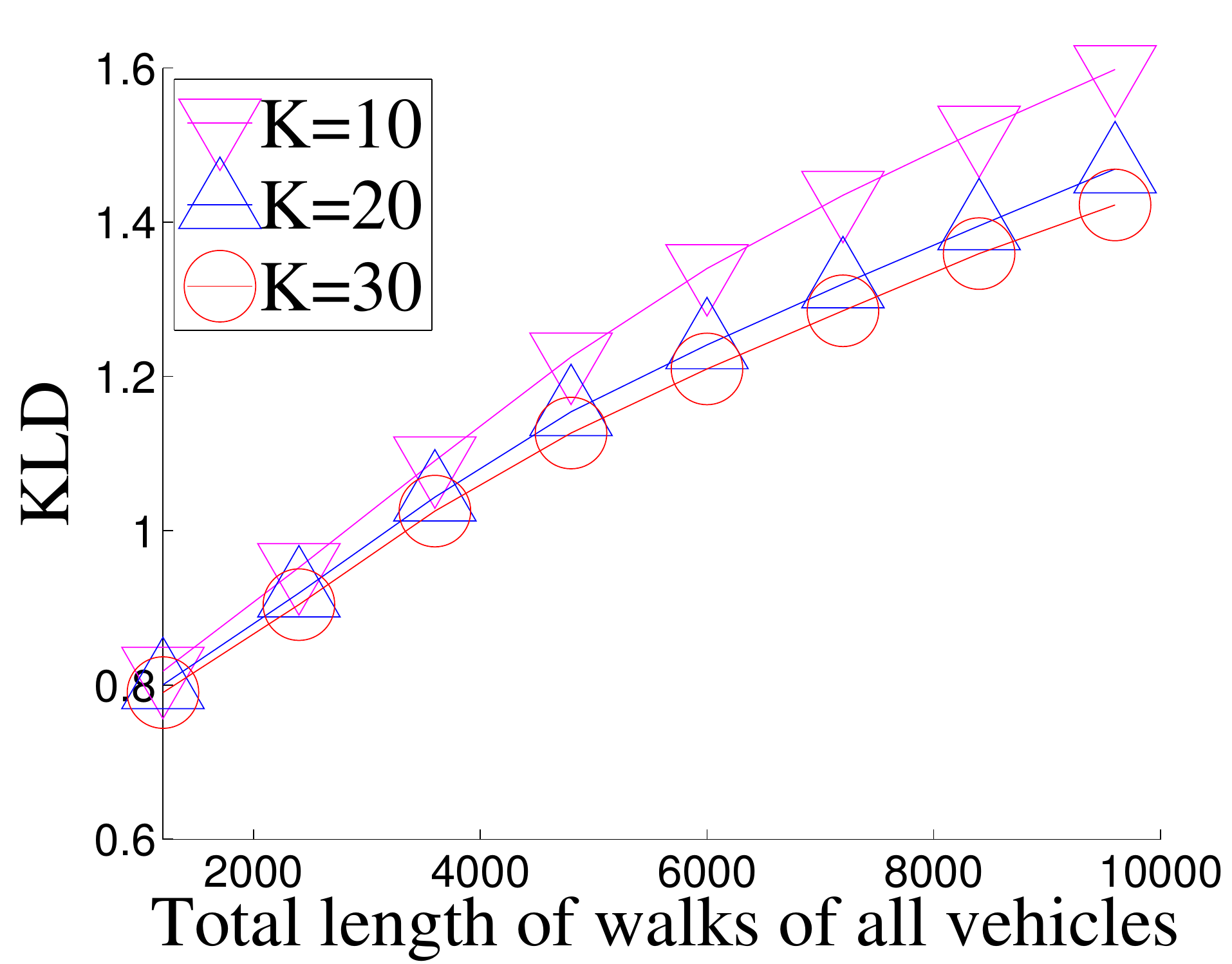}&
\hspace{-5mm} 
\includegraphics[scale=0.15]{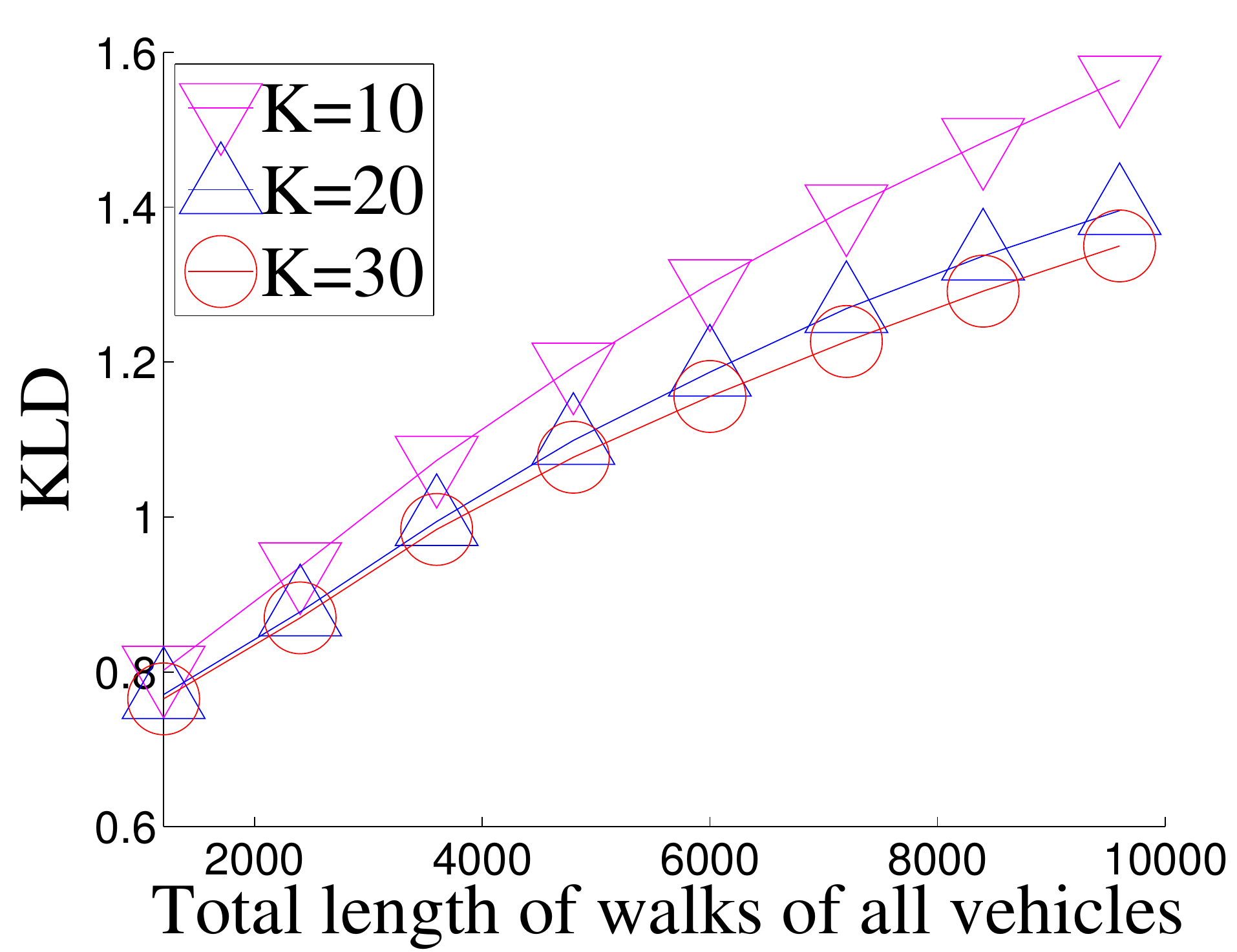}\\
\hspace{-3.5mm} (a) FGP& 
\hspace{-5mm} (b) GP-DDF& 
\hspace{-5mm} (c) GP-DDF$^+$\\
\hspace{-3.5mm}
\includegraphics[scale=0.15]{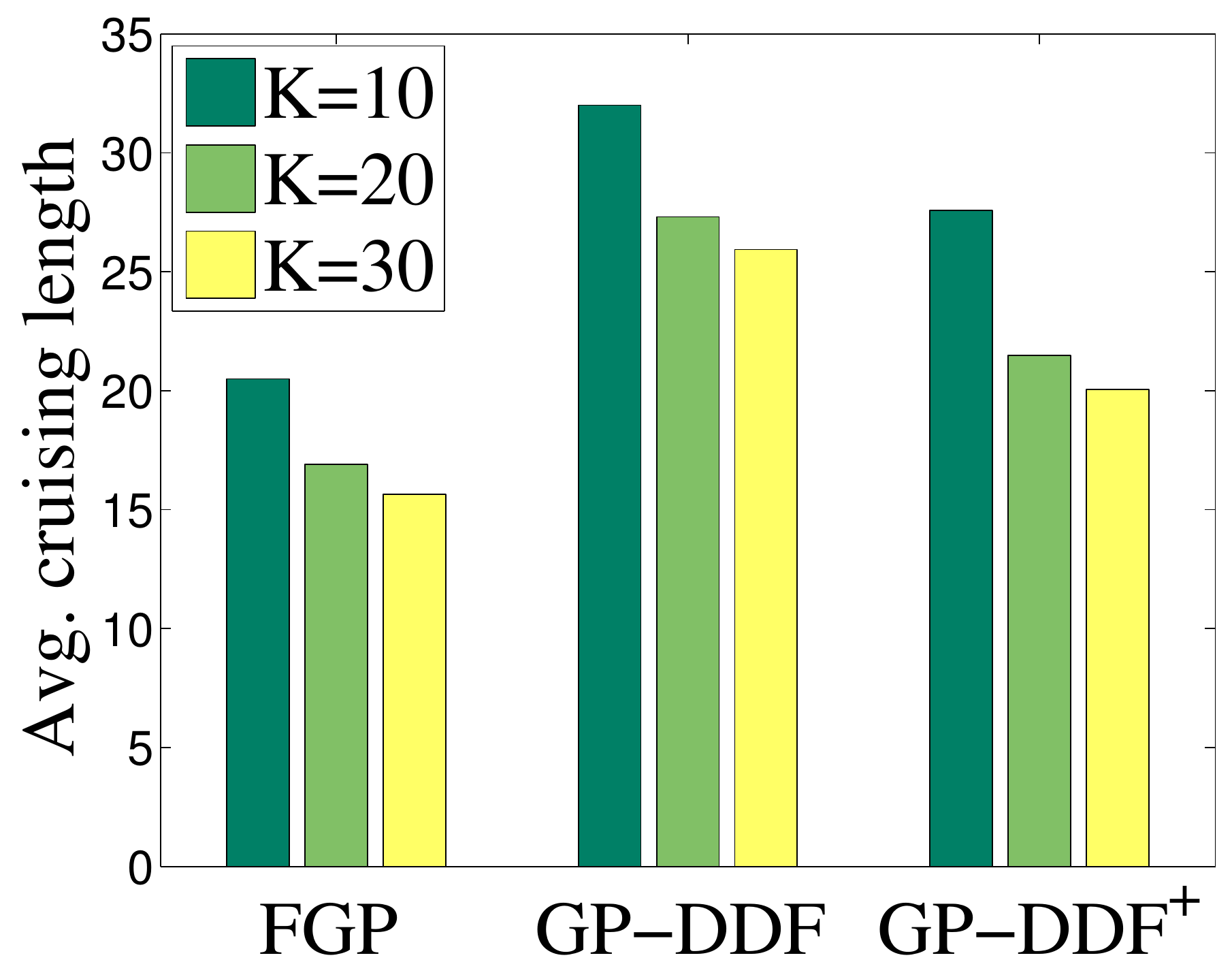}&
\hspace{-5mm} 
\includegraphics[scale=0.15]{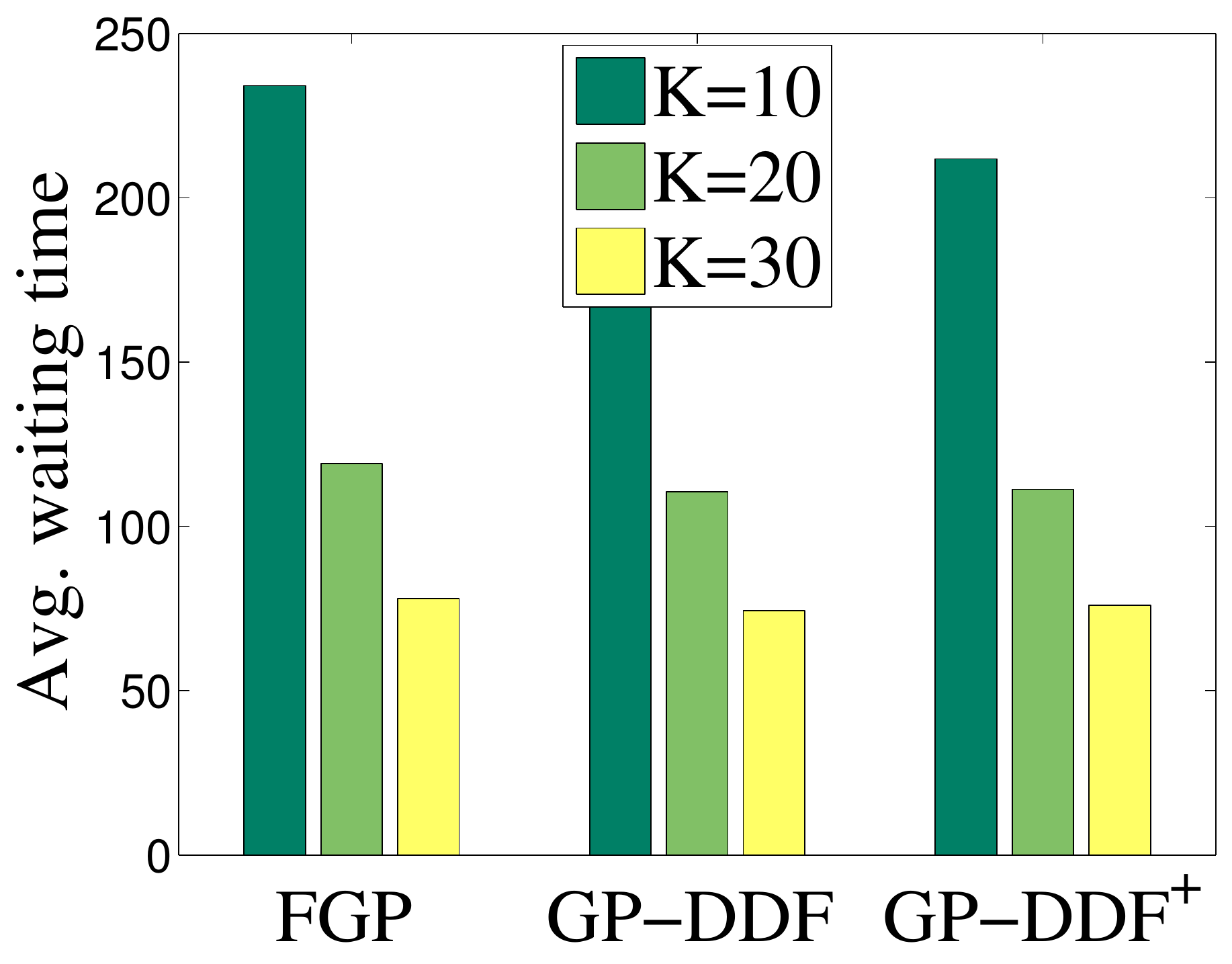}&
\hspace{-5mm} 
\includegraphics[scale=0.15]{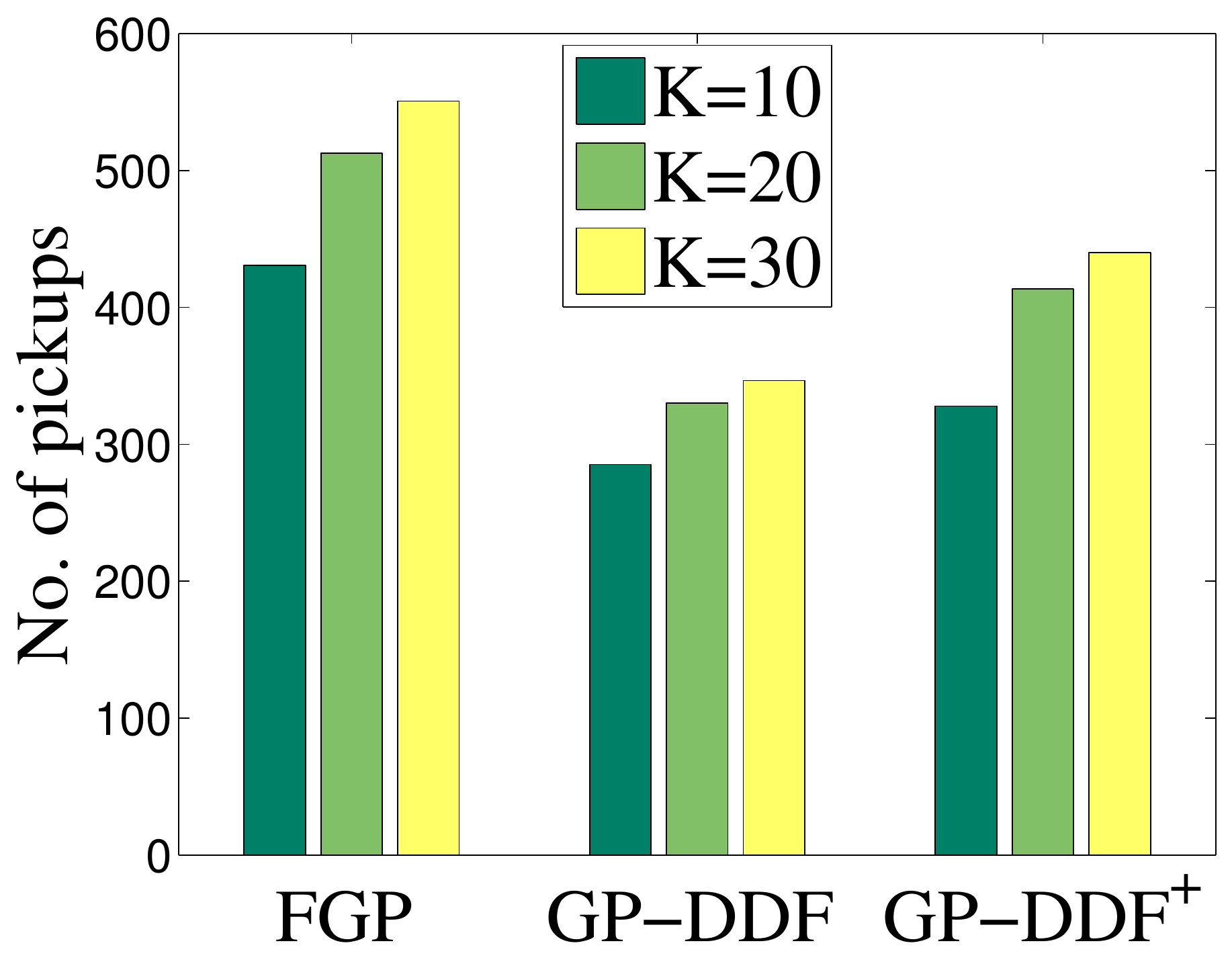}\\ 
\hspace{-3.5mm} (d) Vehicles &
\hspace{-5mm} (e) Users  &
\hspace{-5mm} (f) Pickups \\ 
\end{tabular}
\caption{Scalability of MoD systems in servicing mobility demands.}
\label{fig:rst4} \end{figure}

Figs.~\ref{fig:rst4}a-\ref{fig:rst4}c show that all three algorithms can achieve better
balance between mobility demand and fleet with larger number of
vehicles. It can also be observed that all three algorithms can improve the
performance of servicing the mobility demand with more vehicles, that is, shorter
average cruising trajectories of vehicles (Fig.~\ref{fig:rst4}d),
shorter average waiting time of users (Fig.~\ref{fig:rst4}e),
and larger total number of pickups (Fig.~\ref{fig:rst4}f). This is because MoD
vehicles can collect more informative demand data with larger number of
vehicles sampling demand hotspots or sparsely sampled regions, which are
the regions with higher chance of picking up users than the rest of the service area.

The above results indicate that more vehicles in MoD system result in
better accuracy in predicting the mobility demand pattern, and achieve a
dual effect of better performance in servicing mobility demands.
%

\section{Conclusion} 
This paper describes a novel GP-based decentralized data fusion (GP-DDF$^+$) and active sensing (DAS) algorithm for real-time, fine-grained mobility demand sensing and prediction with a fleet of autonomous robotic vehicles in a MoD system. We have analytically and empirically demonstrated that GP-DDF$^+$ can achieve a better balance between predictive accuracy and time efficiency than the state-of-the-art GP-DDF \cite{chen12} and FGP \cite{LowAAMAS08,low09}.
The practical applicability of GP-DDF$^+$ is not restricted to mobility demand prediction; it can be used in other urban and natural environmental sensing applications like monitoring of traffic, ocean and freshwater phenomena \cite{LowAAMAS13,LowSPIE09,LowICRA07,LowIPSN09,LowAAMAS11,LowAAMAS12,LowAeroconf10,LowICRA13,LowIAT12}.
We have also analytically and empirically shown that even though DAS is devised to gather the most informative demand data for predicting the mobility demand pattern, it can achieve a dual effect of fleet rebalancing to service the mobility demands.
For our future work, we will relax the assumption of all-to-all communication such that each sensor may only be able to communicate locally with its neighbors and develop a distributed data fusion approach to efficient and scalable approximate GP and $\ell$GP prediction.
\section*{Acknowledgments}
This work was supported by Singapore-MIT Alliance Research \& Technology Subaward Agreements No. $28$ R-$252$-$000$-$502$-$592$ \& No. $33$ R-$252$-$000$-$509$-$592$.

\bibliographystyle{plainnat}
\bibliography{mybib}

\if \myproof1
\clearpage
\appendix

\subsection{Proof of Theorem~\ref{thm:ddf}B}
%
%
\vspace{-0mm}\begin{equation}
  \hspace{-0mm} \begin{array}{l} 
\left(\Gamma_{DD}+\Lambda\right)^{-1}\\
\displaystyle =
  \left(\Sigma_{DU}\Sigma_{UU}^{-1}\Sigma_{UD}+\Lambda\right)^{-1}\\
\displaystyle = \Lambda^{-1}-\Lambda^{-1}\Sigma_{DU}
  \left(\Sigma_{UU}+\Sigma_{UD}\Lambda^{-1}\Sigma_{DU}\right)^{-1}
  \Sigma_{UD}\Lambda^{-1}\\ 
    \displaystyle
    =\Lambda^{-1}-\Lambda^{-1}\Sigma_{DU}\ddot{\Sigma}_{UU}^{-1}\Sigma_{UD}\Lambda^{-1}
    \ .
\end{array} \label{pf:pitck} \end{equation}
The first equality is by the definition of $\Gamma_{DD}$ (\ref{eq:sparse}).
The second equality follows from matrix inversion lemma.  The last
equality is due to 
\vspace{-0mm}\begin{equation}
  \begin{array}{l}
\displaystyle\Sigma_{UU}+\Sigma_{UD}\Lambda^{-1}\Sigma_{DU}\\
  =\displaystyle \Sigma_{UU}+\sum_{k=1}^{K}\Sigma_{UD_k}\Sigma_{D_k
    D_k|U}^{-1}\Sigma_{D_kU}\\
  = \displaystyle \Sigma_{UU}+\sum_{k=1}^{K}\dot{\Sigma}_{UU}^k
  =\displaystyle \ddot{\Sigma}_{UU} \ .
\end{array} \label{pf:aa} \end{equation}
The first equality is by the definition of $\Lambda$ (Theorem~\ref{thm:ddf}B). The second and last equalities are due to (\ref{eq:lsc}) and (\ref{eq:gsc}), respectively.

We will first derive the expressions of four components useful for completing the proof later.
\vspace{-0mm}\begin{equation}
\hspace{-1.7mm}
  \begin{array}{l}
\displaystyle
\widetilde{\Gamma}_{sD}\Lambda^{-1} (z_D-\mu_D)\\
\displaystyle
= \sum_{i\neq k}\Gamma_{sD_i}\Sigma_{D_iD_i|U}^{-1}(z_{D_i}-\mu_{D_i})
+\Sigma_{sD_k}\Sigma_{D_kD_k|U}^{-1}(z_{D_k}-\mu_{D_k})\\ 
\displaystyle
= \Sigma_{sU}\Sigma_{UU}^{-1}\sum_{i\neq k}\left(\Sigma_{UD_i}
\Sigma_{D_iD_i|U}^{-1}(z_{D_i}-\mu_{D_i})\right) + \dot{z}_{s}^{k} \\
\displaystyle
= \Sigma_{sU}\Sigma_{UU}^{-1}\sum_{i\neq k}\dot{z}_U^i
+ \dot{z}_{s}^{k} \\
\displaystyle
= \Sigma_{sU}\Sigma_{UU}^{-1}(\ddot{z}_U-\dot{z}_U^k)
+ \dot{z}_{s}^{k} \ .
\end{array}
\label{pf:p1}
\end{equation}
The first two equalities expand the first component using the definition of $\Lambda$ (Theorem~\ref{thm:ddf}B), (\ref{eq:lsf}), (\ref{eq:sparse}), (\ref{eq:pick}), and (\ref{eq:pickernel}). The last two equalities exploit (\ref{eq:lsf}) and
(\ref{eq:gsf}). 
\vspace{-0mm}\begin{equation}
\hspace{-1.7mm}
\begin{array}{l}
\displaystyle
\widetilde{\Gamma}_{sD}\Lambda^{-1} \Sigma_{DU}\\
\displaystyle
= \sum_{i\neq k}\Gamma_{sD_i}\Sigma_{D_iD_i|U}^{-1}\Sigma_{D_iU} 
+\Sigma_{sD_k}\Sigma_{D_kD_k|U}^{-1}\Sigma_{D_kU} \\
\displaystyle
= \Sigma_{sU}\Sigma_{UU}^{-1}\sum_{i\neq k}\left(\Sigma_{UD_i}
\Sigma_{D_iD_i|U}^{-1}\Sigma_{D_iU}\right)
+\Sigma_{sD_k}\Sigma_{D_kD_k|U}^{-1}\Sigma_{D_kU} \\
%
%
\displaystyle
= \Sigma_{sU}\Sigma_{UU}^{-1}\sum_{i\neq k}\dot{\Sigma}_{UU}^i
+\dot{\Sigma}_{sU}^{k} \\
%
\displaystyle
= \Sigma_{sU}\Sigma_{UU}^{-1}\left(\ddot{\Sigma}_{UU}
    -\dot{\Sigma}_{UU}^k-{\Sigma}_{UU}\right)
+\dot{\Sigma}_{sU}^{k} \\
%
%
\displaystyle
=\Sigma_{sU}\Sigma_{UU}^{-1}\ddot{\Sigma}_{UU}-{\gamma}_{sU}^k\ .
\end{array}
\label{pf:p2}
\end{equation}
The first two equalities expand the second component by the same trick as that in
(\ref{pf:p1}). 
The third and fourth equalities exploit (\ref{eq:lsc}) and (\ref{eq:gsc}), respectively. The last equality is due to (\ref{eq:keys}).

Let $\alpha_{sU}\triangleq\Sigma_{sU}\Sigma_{UU}^{-1}$ and its transpose
is $\alpha_{Us}$. By using similar tricks in (\ref{pf:p1}) and
(\ref{pf:p2}), we can derive the expressions of the remaining two components.

If $\tau_s = \tau_{s'}= k$, then
\vspace{-0mm}\begin{equation}
\begin{array}{l}
\displaystyle
\widetilde{\Gamma}_{sD}\Lambda^{-1}\widetilde{\Gamma}_{Ds'}\\
%
\displaystyle
= \sum_{i\neq k}\Gamma_{sD_i}\Sigma_{D_iD_i|U}^{-1}\Gamma_{D_is'} 
+\Sigma_{sD_k}\Sigma_{D_kD_k|U}^{-1}\Sigma_{D_ks'} \\
\displaystyle
= \Sigma_{sU}\Sigma_{UU}^{-1}\sum_{i\neq k}\left(\Sigma_{UD_i}
\Sigma_{D_iD_i|U}^{-1}\Sigma_{D_iU}\right)\Sigma_{UU}^{-1}\Sigma_{Us'}+\dot{\Sigma}_{ss'}^{k} \\
\displaystyle
= \alpha_{sU}\sum_{i\neq k}\left(
    \dot{\Sigma}_{UU}^i
\right)\alpha_{Us'}
+ \dot{\Sigma}_{ss'}^{k} \\
\displaystyle
= \alpha_{sU}\left(\ddot{\Sigma}_{UU}-\dot{\Sigma}_{UU}^k-{\Sigma}_{UU}
\right)\alpha_{Us'} + \dot{\Sigma}_{ss'}^{k} \\
\displaystyle
=\alpha_{sU}\ddot{\Sigma}_{UU}\alpha_{Us'}
    -\alpha_{sU}{\gamma}_{Us'}^k - \alpha_{sU}\dot{\Sigma}^k_{Us'} 
+ \dot{\Sigma}_{ss'}^{k} \ .
\end{array}
\label{pf:p3}
\end{equation}
If $\tau_s = i$ and $\tau_{s'}= j$ such that $i\neq j$, then
\vspace{-0mm}\begin{equation}
\begin{array}{l}
\displaystyle
\widetilde{\Gamma}_{sD}\Lambda^{-1}\widetilde{\Gamma}_{Ds'}\\
\displaystyle
=\sum_{k\neq i,j}\Gamma_{sD_k}\Sigma_{D_kD_k|U}^{-1}\Gamma_{D_ks'}\\ 
\quad +\ \Sigma_{sD_i}\Sigma_{D_iD_i|U}^{-1}\Gamma_{D_is'}
 +\Gamma_{sD_j}\Sigma_{D_jD_j|U}^{-1}\Sigma_{D_js'}\\ 
\displaystyle
=\Sigma_{sU}\Sigma_{UU}^{-1}\sum_{k\neq
i,j}\left(\Sigma_{UD_k}\Sigma_{D_kD_k|U}^{-1}\Sigma_{D_kU}\right)
\Sigma_{UU}^{-1}\Sigma_{Us'}\\ 
\quad +\ \Sigma_{sD_i}\Sigma_{D_iD_i|U}^{-1}\Sigma_{D_iU}\Sigma_{UU}^{-1}\Sigma_{Us'}
 +\Sigma_{sU}\Sigma_{UU}^{-1}\Sigma_{UD_j}\Sigma_{D_jD_j|U}^{-1}\Sigma_{D_js'}\\ 
%
%
\displaystyle
=\alpha_{sU}\left(\ddot{\Sigma}_{UU}-\dot{\Sigma}_{UU}^i
  -\dot{\Sigma}_{UU}^j-{\Sigma}_{UU}\right)\alpha_{Us'} + \dot{\Sigma}^i_{sU}\alpha_{Us'}
 +{\alpha}_{sU}\dot{\Sigma}^j_{Us'}\\ 
\displaystyle
=\alpha_{sU}\left(\ddot{\Sigma}_{UU}+{\Sigma}_{UU}\right)\alpha_{Us'} 
 -\alpha_{sU}\left(\dot{\Sigma}_{UU}^i+{\Sigma}_{UU}\right)\alpha_{Us'}\\
\quad-\ \alpha_{sU}\left(\dot{\Sigma}_{UU}^j+{\Sigma}_{UU}\right)\alpha_{Us'} 
+ \dot{\Sigma}^i_{sU}\alpha_{Us'}
 +{\alpha}_{sU}\dot{\Sigma}^j_{Us'}\\ 
%
%
\displaystyle
= \alpha_{sU}\left(\ddot{\Sigma}_{UU}+{\Sigma}_{UU}\right)\alpha_{Us'}
- \left(\alpha_{sU}\dot{\Sigma}_{UU}^i+\alpha_{sU}{\Sigma}_{UU} - \dot{\Sigma}^i_{sU}\right)\alpha_{Us'}\\
\quad -\ \alpha_{sU}\left(\dot{\Sigma}_{UU}^j \alpha_{Us'}+{\Sigma}_{UU}\alpha_{Us'}-\dot{\Sigma}^j_{Us'}\right)\\ 
\displaystyle
  =\alpha_{sU}\left(\ddot{\Sigma}_{UU}+{\Sigma}_{UU}\right)\alpha_{Us'}
-{\gamma}_{sU}^i \alpha_{Us'} -{\alpha}_{sU} {\gamma}_{Us'}^j\\
\displaystyle
  =\alpha_{sU}\ddot{\Sigma}_{UU}\alpha_{Us'}
  +\Sigma_{sU}{\Sigma}_{UU}^{-1}\Sigma_{Us'}
-{\gamma}_{sU}^i \alpha_{Us'} -{\alpha}_{sU} {\gamma}_{Us'}^j\\
  \end{array}
  \label{pf:p4}
\end{equation}

If $\tau_s = k$, then
$$
\begin{array}{l}
\mu^{\mbox{\tiny{PIC}}}_{s|D}\\
= \mu_s +
\widetilde{\Gamma}_{sD}\left(\Gamma_{DD}+\Lambda\right)^{-1}(z_D-\mu_D)\\
=\mu_s + \widetilde{\Gamma}_{sD}\Lambda^{-1}(z_D-\mu_D)\\
\quad -\ \widetilde{\Gamma}_{sD}\Lambda^{-1}\Sigma_{DU}\ddot{\Sigma}_{UU}^{-1}\Sigma_{UD}\Lambda^{-1}(z_D-\mu_D)\\
=\mu_s + \widetilde{\Gamma}_{sD}\Lambda^{-1}(z_D-\mu_D)
- \widetilde{\Gamma}_{sD}\Lambda^{-1}\Sigma_{DU}\ddot{\Sigma}_{UU}^{-1}\ddot{z}_U\\
=\mu_s + \Sigma_{sU}\Sigma_{UU}^{-1}(\ddot{z}_U-\dot{z}_U^k)
+\dot{z}_{s}^{k} - \widetilde{\Gamma}_{sD}\Lambda^{-1}\Sigma_{DU}\ddot{\Sigma}_{UU}^{-1}\ddot{z}_U\\
=\mu_s + \Sigma_{sU}\Sigma_{UU}^{-1}(\ddot{z}_U-\dot{z}_U^k)
+\dot{z}_{s}^{k} \\
\quad-\left(\Sigma_{sU}\Sigma_{UU}^{-1}\ddot{\Sigma}_{UU}-{\gamma}_{sU}^k
\right) \ddot{\Sigma}_{UU}^{-1}\ddot{z}_U\\
=\mu_s +\left({\gamma}_{sU}^k\ddot{\Sigma}_{UU}^{-1}\ddot{z}_U
    -\Sigma_{sU}\Sigma_{UU}^{-1}\dot{z}_{U}^k\right) +\dot{z}_{s}^{k}\\
=  \overline{\mu}_s^{k}\ .  
\end{array}
$$
The first equality is by definition (\ref{eq:picmu}).
The second equality is due to (\ref{pf:pitck}). The third equality is
due to the definition of global summary (\ref{eq:gsf}). The fourth and fifth equalities
are due to (\ref{pf:p1}) and (\ref{pf:p2}), respectively.

If $\tau_s = \tau_{s'}= k$, then
$$
\begin{array}{l}
\sigma_{ss'|D}^{\mbox{\tiny{PIC}}}\\
= \sigma_{ss'}- 
\widetilde{\Gamma}_{sD}\left(\Gamma_{DD}+\Lambda\right)^{-1}\widetilde{\Gamma}_{Ds'}\\
= \sigma_{ss'}-\widetilde{\Gamma}_{sD}\Lambda^{-1}\widetilde{\Gamma}_{Ds'}
+\widetilde{\Gamma}_{sD}\Lambda^{-1}\Sigma_{DU}\ddot{\Sigma}_{UU}^{-1}
\Sigma_{UD}\Lambda^{-1}\widetilde{\Gamma}_{Ds'}\\
=\sigma_{ss'}
-\widetilde{\Gamma}_{sD}\Lambda^{-1}\widetilde{\Gamma}_{Ds'}\\
\quad+\left(\alpha_{sU}\ddot{\Sigma}_{UU}-{\gamma}_{sU}^k\right)
\ddot{\Sigma}_{UU}^{-1}
\left(\ddot{\Sigma}_{UU}\alpha_{Us'}-{\gamma}_{Us'}^k\right)\\
%
=\sigma_{ss'} -
\alpha_{sU}\ddot{\Sigma}_{UU}\alpha_{Us'}
    +\alpha_{sU}{\gamma}_{Us'}^k +\alpha_{sU}\dot{\Sigma}^k_{Us'} - \dot{\Sigma}_{ss'}^{k}\\ 
\quad+\ \alpha_{sU}\ddot{\Sigma}_{UU}\alpha_{Us'}
    -\alpha_{sU}\gamma_{Us'}^k -\gamma_{sU}^k\alpha_{Us'}+{\gamma}_{sU}^k\ddot{\Sigma}_{UU}^{-1}{\gamma}_{Us'}^k \\
%
%
=\sigma_{ss'} -\left(\gamma_{sU}^k\alpha_{Us'} -\alpha_{sU}\dot{\Sigma}^k_{Us'} -{\gamma}_{sU}^k\ddot{\Sigma}_{UU}^{-1}{\gamma}_{Us'}^k\right)- \dot{\Sigma}_{ss'}^{k}\\
= \overline{\sigma}_{ss'}^{k}\\
= \overline{\sigma}_{ss'}\ .
%
%
\end{array}
$$
The first equality is by definition (\ref{eq:picvar}).
The second equality is due to (\ref{pf:pitck}). The third equality is
due to (\ref{pf:p2}). The fourth equality is due to (\ref{pf:p3}). The
last two equalities are by definition (\ref{eq:dpiccov}).

If $\tau_s = i$ and $\tau_{s'}= j$ such that $i\neq j$, then
$$
\begin{array}{l}
\sigma_{ss'|D}^{\mbox{\tiny{PIC}}}\\
=\sigma_{ss'}
-\widetilde{\Gamma}_{sD}\Lambda^{-1}\widetilde{\Gamma}_{Ds'}\\
\quad +\ \alpha_{sU}\ddot{\Sigma}_{UU}\alpha_{Us'}
-\alpha_{sU}\gamma_{Us'}^j     -\gamma_{sU}^i\alpha_{Us'}
    +{\gamma}_{sU}^i\ddot{\Sigma}_{UU}^{-1}{\gamma}_{Us'}^j \\
=\sigma_{ss'}-\left(\alpha_{sU}\ddot{\Sigma}_{UU}\alpha_{Us'}
+\Sigma_{sU}{\Sigma}_{UU}^{-1}\Sigma_{Us'}
-{\gamma}_{sU}^i \alpha_{Us'} -{\alpha}_{sU} {\gamma}_{Us'}^j
  \right)\\
\quad+\ \alpha_{sU}\ddot{\Sigma}_{UU}\alpha_{Us'}
-\alpha_{sU}\gamma_{Us'}^j-\gamma_{sU}^i\alpha_{Us'}
+{\gamma}_{sU}^i\ddot{\Sigma}_{UU}^{-1}{\gamma}_{Us'}^j\\
=\sigma_{ss'} -\Sigma_{sU}{\Sigma}_{UU}^{-1}\Sigma_{Us'}
+{\gamma}_{sU}^i\ddot{\Sigma}_{UU}^{-1}{\gamma}_{Us'}^j\\
=\Sigma_{ss'|U}+{\gamma}_{sU}^i\ddot{\Sigma}_{UU}^{-1}{\gamma}_{Us'}^j\\
= \overline{\sigma}_{ss'}\ .
\end{array}
$$
The first equality is obtained using a similar trick as when $\tau_s = \tau_{s'}= k$. The second equality is due to (\ref{pf:p4}). The second last
equality is by the definition of posterior covariance in GP model (\ref{eq:fgpcov}).
The last equality is by definition (\ref{eq:dpiccov}).
\fi
\end{document}